\documentclass{article}

\usepackage{microtype}
\usepackage{graphicx}
\usepackage{subcaption}
\usepackage{booktabs} 
\usepackage{bm}
\usepackage{hyperref}

\usepackage[noend]{algorithmic}
\renewcommand{\algorithmicrequire}{\textbf{Input:}}
\renewcommand{\algorithmicensure}{\textbf{Output:}}

\usepackage[accepted]{icml2025}

\usepackage{amsmath}
\usepackage{amssymb}
\usepackage{mathtools}
\usepackage{amsthm}
\usepackage{enumitem}
\usepackage[capitalize,noabbrev]{cleveref}

\newcommand{\nate}[1]{{\color{black}{#1}}}

\newcommand{\doc}[1]{{\color{teal}{#1}}}

\usepackage{listings}
\usepackage{tcolorbox}

\definecolor{dkgreen}{rgb}{0,0.6,0}
\definecolor{gray}{rgb}{0.5,0.5,0.5}
\definecolor{mauve}{rgb}{0.58,0,0.82}

\theoremstyle{plain}

\theoremstyle{definition}
\newtheorem{definition}{Definition}

\theoremstyle{remark}

\usepackage[textsize=tiny]{todonotes}

\icmltitlerunning{Large Language Models for Multi-Facility Location Mechanism Design}

\begin{document}

\twocolumn[
\icmltitle{Large Language Models for Multi-Facility Location Mechanism Design}




\begin{icmlauthorlist}
\icmlauthor{Nguyen Thach}{unl}
\icmlauthor{Fei Liu}{cityuhk}
\icmlauthor{Houyu Zhou}{unsw}
\icmlauthor{Hau Chan}{unl}
\end{icmlauthorlist}

\icmlaffiliation{unl}{University of Nebraska-Lincoln}
\icmlaffiliation{cityuhk}{City University of Hong Kong}
\icmlaffiliation{unsw}{UNSW Sydney}

\icmlcorrespondingauthor{Nguyen (Nate) Thach}{nate.thach@huskers.unl.edu}
\icmlcorrespondingauthor{Fei Liu}{fliu36-c@my.cityu.edu.hk}
\icmlcorrespondingauthor{Houyu Zhou}{houyu.zhou@unsw.edu.au}
\icmlcorrespondingauthor{Hau Chan}{hchan3@unl.edu}

\icmlkeywords{Machine Learning, ICML}

\vskip 0.3in
]



\printAffiliationsAndNotice{}  

\begin{abstract}

Designing strategyproof mechanisms for multi-facility location that optimize social costs based on agent preferences had been challenging due to the extensive domain knowledge required and poor worst-case guarantees. Recently, deep learning models have been proposed as alternatives. 
However, these models require some domain knowledge and extensive hyperparameter tuning as well as lacking interpretability, which is crucial in practice when transparency of the learned mechanisms is mandatory. 
In this paper, we introduce a novel approach, named LLMMech, that addresses these limitations by incorporating large language models (LLMs) into an evolutionary framework for generating interpretable, hyperparameter-free, empirically strategyproof, and nearly optimal mechanisms. 
Our experimental results, evaluated on various problem settings where the social cost is arbitrarily weighted across agents and the agent preferences may not be uniformly distributed, demonstrate that the LLM-generated mechanisms generally outperform existing handcrafted baselines and deep learning models. Furthermore, the mechanisms exhibit impressive generalizability to out-of-distribution agent preferences
and to larger instances with more agents. 


\end{abstract}

\section{Introduction}

In recent decades, mechanism design for facility location has garnered significant interest from various communities, including operations research, economics, artificial intelligence, and machine learning. 
This is due to their ability to inform collective decision-making, e.g., in healthcare, urban planning, and preference aggregation applications \cite{Chan:2021aa,Ahmadi-Javid:2017aa,Black:1948aa,moulin1980strategy,Farahani:2009aa} based on (individual) agent preferences. 
In a standard mechanism design setting for facility location, the social planner seeks to locate several facilities (e.g., parks, transit stations, or clinics) to serve agents within a region by considering agent preferences on the ideal locations of the facilities and optimizing social costs based on agent preferences (e.g., the total distances between agent ideal locations and facility locations). 
Unfortunately, it is well documented that agents have the incentive to misrepresent or misreport preferences to manipulate facility locations, which can lead to undesirable societal situations.  
Therefore, existing related studies have proposed to design \emph{strategyproof} mechanisms that elicit agent preferences truthfully and determine facility locations to (approximately) optimize a given social cost.

\paragraph{Existing Efforts.} 
The mechanism design studies for facility location were first examined by \citet{moulin1980strategy} and \citet{procaccia2013approximate}, who characterized the set of strategyproof mechanisms for locating one facility and developed strategyproof mechanisms (with worst-case theoretical approximation guarantees) for locating one or two facilities to optimize a given social cost when facility locations and agent preferences are on the real line, respectively. 
Following these studies, existing efforts have focused on designing strategyproof mechanisms for various facility location variants (e.g., considering fairness/envy objectives, higher dimensional space, facilities with constraints, and other types of agent preferences) based on two main technical approaches \cite{Chan:2021aa}. 
In the first (most common) technical approach, existing efforts aimed to design mechanisms manually and show theoretically that they are both strategyproof and have approximation guarantees through domain knowledge and handcrafting designs \cite{Chan:2021aa}. 
Because the first approach relies on extensive domain expertise and often yields mechanisms that have poor worst-case guarantees, \cite{golowich2018deep} proposed a second technical approach to use deep learning to design mechanisms (e.g., via neural networks) that are approximately strategyproof and optimal empirically. 

However, the second approach requires some domain knowledge and extensive hyperparameter tuning of the neural network architectures to design mechanisms, which might not be available in practice and may not return (approximately) strategyproof mechanisms even when carefully tuned, respectively. 
Moreover, their designed mechanisms are far from being interpretable, which is crucial for explaining the mechanisms to the social planner and helping the mechanism designers gain additional insight into discovering new mechanisms. 
Therefore, in this paper, we strive to address the following question. 
\vspace{-5pt}
\begin{quote}
\emph{Can we design mechanisms that are hyperparameter-free, interpretable, approximately strategyproof, and optimal empirically for optimizing a given objective?}
\end{quote}

\paragraph{Our Approaches and Contributions.} 
Large language models (LLMs) have shown remarkable capabilities in understanding and generating human-like text \cite{brown2020language}, solving complex optimization problems \cite{liu2024systematic}, and even programming \cite{jiang2024survey}. 
Therefore, to address the above question affirmatively, we propose a new approach that leverages LLMs in this paper. 
We summarize our contributions as follows:






\begin{itemize}[itemsep=-2pt,topsep=-2pt]
\item We introduce a novel framework, named LLMMech, that integrates an LLM into an evolutionary framework to automate the design of empirically strategyproof and nearly optimal mechanisms for locating multiple facilities. 
\nate{LLMMech addresses the limitations of existing approaches by enabling the automated creation of customized and fully interpretable heuristic mechanisms while requiring minimal domain knowledge and hyperparameter tuning, which significantly reduces the manual effort involved in mechanism design and potentially brings fresh insights to the problem at hand.}

\item 
\nate{We extend the framework of EoH \cite{liu2023algorithm,liu2024evolution} and facilitate the evolutionary search of novel mechanisms by incorporating automatic prompt evolution within LLMMech whenever the search process stagnates and seemingly converges to a local optimum (within the landscape defined by the social cost). This extension, in turn, reduces the efforts needed for manually designing prompt strategies, which guide the LLM to reason over existing mechanisms \cite{liu2024evolution}.}

\item  We empirically demonstrate that LLMMech \nate{outperforms existing methods, including the state-of-the-art deep learning models \cite{golowich2018deep}, on the general classical settings \cite{moulin1980strategy,procaccia2013approximate} where the social cost is arbitrarily weighted across agents and the agent preferences may not be uniformly distributed. 
Furthermore, LLMMech generates mechanisms that can generalize to out-of-distribution agent preferences and to larger instances with more agents than what was originally learned.} 
\end{itemize}


\paragraph{Outline.}
In Section \ref{sec:pre}, we provide preliminaries for our considered facility location problem.
In Section \ref{sec:method}, we elaborate the LLMMech framework.
In Section \ref{sec:exp}, we evaluate its efficacy through various experiments.
We conclude our work in Section \ref{sec:conclusion}.
Readers can find related work and missing experimental details/results in the Appendices.

\section{Preliminaries}
\label{sec:pre}

In this section, we begin by presenting the general classical settings of \cite{moulin1980strategy,procaccia2013approximate} that were considered in the mechanism design learning problem of \cite{golowich2018deep}.
More specially, there are a set of agents $\mathcal{N} = \{1, \dots, n\}$ and a set of locations $\Omega = [0,1]$ specifying the domain where facilities can be placed. 
The social planner is interested in the problem of locating $K$ facilities among locations in $\Omega$. 
Each agent $i \in N$ has a preference over $\Omega$ for where a facility should be located, represented by a cost function $u_i : \Omega \to \mathbb{R}$ that associates a real value with each location. 
Let $U_i$ be the set of permissible cost functions. Let $u = (u_1, \dots, u_n)$ denote a profile of costs, and denote the set of all such tuples by $U = \prod_{i=1}^n U_i$. Further, let $u_{-i} = (u_1, \dots, u_{i-1}, u_{i+1}, \dots, u_n)$ denote all cost functions other than $u_i$, and $U_{-i} = \prod_{j \neq i} U_j$.

Each agent cares about the location of its closest facility. As such, an agent’s cost function induces a cost for an outcome, $o = (o_1, \dots, o_K)$, which specifies a location for each facility. By a slight abuse of notation, we write $u_i(o) = \min_{k \in \{1, \dots, K\}} u_i(o_k)$. 
In existing studies, the agents have \emph{single-peaked} preferences.

\begin{definition}[Single-peaked preferences]
    For locations $\Omega \subseteq \mathbb{R}$, a function $u_i : \Omega \to \mathbb{R}_{\geq 0}$ is \textit{single-peaked} if and only if there exists a unique point $a \in \Omega$ (the \textit{peak} of $u_i$, denoted by $\tau(u_i)$) such that for all $x, y \in \Omega$:
    $\text{if } x > a \text{ and } y > x, \text{ then } u_i(y) > u_i(x),$
    $\text{and if } x < a \text{ and } y < x, \text{ then } u_i(y) > u_i(x).$
\end{definition}

As one moves away from the peak $\tau(u_i)$ of cost function $u_i$, the location becomes less preferred. Following \cite{moulin1980strategy,procaccia2013approximate,golowich2018deep} and subsequent work, the cost is defined as $u_i(x) = |x - a|$, where $a = \tau(u_i)$ is the peak. 



A \textit{mechanism} for facility location $f : U \to \Omega^K$ takes reports of agent cost functions as inputs and outputs the locations in $\Omega$ for the $K$ facilities.
Since agent preferences are single-peaked, with a slight abuse of notation, we instead let $f$ takes agent peaks as inputs for convenience.
We use $f_k(u)$ to denote the location of the $k$-th facility for input $u$. A mechanism is \textit{strategyproof} if agent $i$ cannot strictly decrease their cost by misreporting their cost function, whatever the inputs from others $u_{-i}$, i.e., for all $i \in N$, all $u_i, u_i' \in U_i$, we have $u_i(f(u_i', u_{-i})) \geq u_i(f(u))$. 

We denote by $\mathcal{M}_{sp} \subseteq \mathcal{M}$ the space of all strategyproof mechanisms for facility location (leaving $K$ implicit).
In accordance with \cite{golowich2018deep}, the agent cost functions are sampled from a joint distribution $\mathcal{D}$ over $U$, with full support on $U$. 
Moreover, $\mathcal{D}$ is a product distribution, i.e., $\mathcal{D} = \prod_{i=1}^n \mathcal{D}_i$ for independent distributions $\mathcal{D}_i$ on $U_i$. 

The goal is to find a strategyproof mechanism $f \in \mathcal{M}_{sp}$ that minimizes the expected (weighted) social cost: 
\[
g_{sc}(f; \mathcal{D}) = \mathbb{E}_{u \sim \mathcal{D}}\left[\frac{1}{\sum^n_{i=1}\gamma_i} \sum_{i=1}^n \gamma_iu_i(f(u))\right],
\]
where $\gamma_i$ is the weight associated with agent $i$ and is determined by the social planner who weighs the importance of their preference relative to other agents. 
We do not have direct access to $\mathcal{D}$, but are provided a sample of profiles $S = \{u^{(1)}, \dots, u^{(R)}\}$ drawn i.i.d. from $\mathcal{D}$. 
The problem is then to learn the explicit form of $f$ given the profile sample.
We define a \emph{problem instance} as an instantiation of the problem, and 
a \emph{problem setting} as a set of problem instances that share the same configuration of $n$, $K$, $\mathcal{D}$, and $\{\gamma_i\}$.



\section{LLMs for Mechanism Design (LLMMech)}
\label{sec:method}

Algorithm \ref{alg:lds} provides an overview of our framework. Inspired by EoH \cite{liu2023algorithm,liu2024evolution}, LLMMech maintains a population of $N$ heuristic mechanisms, denoted as $\bm{M} = \{f_1,\ldots,f_N\}$. It adopts an evolutionary search framework, iteratively searching for mechanisms that have lower social costs. 
The mechanisms are designed by prompting pre-trained LLMs. 
Each mechanism $f_l\in \bm{M}$ is evaluated on a set of problem instances and assigned a fitness value $q(f_l)$. We propose an additional high-level prompt evolution procedure to automatically evolve prompts instead of sticking to hand-crafted prompts to avoid premature convergence at local optimal mechanisms. The prompt evolution component is marked in teal, with its main subroutine `PromptEvolution' described in Algorithm \ref{alg:evolop}.
The remainder of this section is organized as follows: Section \ref{subsec:mech-rep} describes the components we used in LLMMech to represent a mechanism; Section \ref{subsec:evol-components} elaborates the abstracted functions in Algorithms \ref{alg:lds} and \ref{alg:evolop}; Section \ref{subsec:fitness} defines the fitness function for evaluating individual mechanisms in a population; Section \ref{subsec:prompt-evo} details the prompt evolution process.



\begin{algorithm}[tb]
    \footnotesize
   \caption{LLMMech}
   \label{alg:lds}
\begin{algorithmic}[1]
   \REQUIRE Pre-trained LLM $\mathcal{L}$; Population size $N$; Evaluation budget $T_e$; Variation prompt set $\bm{P}$; \doc{Prompt population size $N_p$; Patience $T_p<T_e$; Positive integer $b<N$}.
   \ENSURE Mechanism $f^*\in \bm{M}$ with best fitness value.
   \STATE $\bm{M} \leftarrow \varnothing$
   \WHILE{$|\bm{M}|<N$}
   \STATE $M \leftarrow \textit{InitializationPrompt}()$
   \ENDWHILE
   \STATE $t\leftarrow |\bm{M}|$
   \STATE \doc{$\bm{M}^*\leftarrow\textit{Select}(\bm{M},b)$} \hfill\doc{\% $|\bm{M}^*|=b$}
   \STATE \doc{$t'\leftarrow0$}
   \REPEAT
   \STATE $\bm{M}^\prime \leftarrow \varnothing$
   \FOR{$i=1, 2, \ldots, N$}
   \STATE $M_i \leftarrow \textit{Select}(\bm{M})$
   \STATE $M_i^\prime\leftarrow 
   \textit{VariationPrompt}(\mathcal{L},M_i,\bm{P})$
   \STATE $\bm{M}^\prime \leftarrow \bm{M}^\prime \cup M_i^\prime$
   \ENDFOR
   \STATE $\bm{M} \leftarrow \textit{PopulationManager}(\bm{M}, \bm{M}^\prime, N)$
   \STATE $t\leftarrow t + |\bm{M}^\prime|$
   \STATE \doc{$\bm{M}^*_{temp}\leftarrow\textit{Select}(\bm{M},b)$}
   \doc{\IF{$\bm{M}^*_{temp}=\bm{M}^*$}
       \STATE $t'\leftarrow t'+ |\bm{M}^\prime|$
    \ELSE
        \STATE $\bm{M}^*\leftarrow\bm{M}^*_{temp}$
        \STATE $t'\leftarrow0$
   \ENDIF}
   \doc{\IF{$t'\ge T_p$}
       \STATE $\bm{P}\leftarrow\text{\hyperref[alg:evolop]{PromptEvolution}}(\mathcal{L},\bm{P},N_p)$
    \ENDIF}
   \UNTIL{$t\geq T_e$}
\end{algorithmic}
\end{algorithm}

\begin{algorithm}[tb]
    \footnotesize
   \caption{PromptEvolution}
   \label{alg:evolop}
\begin{algorithmic}[1]
   \REQUIRE Pre-trained LLM $\mathcal{L}$; Variation prompt set $\bm{P}$; Prompt population size $N_p$.
   \ENSURE Updated $\bm{P}$.
    \STATE $\bm{P}^\prime \leftarrow \varnothing$
    \FOR{$j=1,2,\ldots,|\bm{P}|$}
        \STATE $P_j \leftarrow \textit{Select}(\bm{P})$
        \STATE $P_j^\prime\leftarrow 
        \textit{EvolveVariationPrompt}(\mathcal{L},P_j)$
        \STATE $\bm{P}^\prime \leftarrow \bm{P}^\prime \cup P_j^\prime$
    \ENDFOR
    \STATE $\bm{P} \leftarrow \textit{PopulationManager}(\bm{P}, \bm{P}^\prime, N_p)$
\end{algorithmic}
\end{algorithm}

\subsection{Mechanism Representation}\label{subsec:mech-rep}

Each mechanism consists of three parts: 


1) The mechanism description comprises a few sentences in natural language. It is created by LLMs and encapsulates a high-level thought. An example is provided in Figure \ref{fig:learned-mech}. 

2) The code block is an implementation of the mechanism. It should follow a pre-defined format so that it can be identified and seamlessly integrated into the evolutionary framework.
In the experiments, we choose to implement it as a Python function\footnote{The function can be written in any programming language.} (see Appendix \ref{app:code-template} for the code templates used). 

3) Each mechanism is assigned a fitness value to represent its priority in the population, which is used for selection and population management. We elaborate our definition of fitness in Section \ref{subsec:fitness}.

\subsection{Evolutionary Components}\label{subsec:evol-components}

We start with the design of prompt strategies.
\nate{Note that we refer to `prompt' as the entire natural-language input to LLMs, and `prompt strategy' as part of the prompt that provides specific instructions for a certain evolutionary operation (e.g., crossover and mutation).} 
Details of the exact prompts used are illustrated in 
Appendix \ref{subapp:var-prompt}.


\paragraph{Initialization Prompt.}
In the \textit{InitializationPrompt} function, we inform the LLM of the facility location of interest and instruct it to design a new mechanism by first presenting the description of the mechanism and the corresponding Python code block (see Figure \ref{fig:init_prompt-design}, top, for an example). We repeat $N$ times to generate $N$ initial mechanisms.

\paragraph{Variation Prompts.}
We design prompt strategies that either \emph{explore} or \emph{modify} existing mechanisms for generating offsprings.
The exploration strategies focus more on the exploration of the space of heuristics by conducting crossover-like operators on parent heuristics. The modification strategies refine a parent heuristic by modifying, adjusting parameters, and removing redundant parts.
With a slight abuse of notation, the \textit{VariationPrompt} function denotes the generation of offspring mechanisms by prompting the LLM with parent mechanism(s) $M_i$ and either exploration or modification prompt strategy in $\bm{P}$ as input.
We only need to provide \emph{one} (instead of two to three) prompt strategy for each type since they are also part of the evolutionary process (to be discussed in Section \ref{subsec:prompt-evo}).
This alleviates the need for extensive prompt engineering, which often requires some degree of domain knowledge especially when dealing with complex problems \cite{liu2024systematic}.
In our case, we design the modification prompt strategy in accordance to our definition of fitness (to be elaborated in Section \ref{subsec:fitness}), which only requires basic understandings of the facility location problem, particularly the definition of `strategyproofness'.
Please refer to Figure \ref{fig:init_prompt-design} (bottom right) from Appendix \ref{subapp:var-prompt} for exact details.

\paragraph{Selection and Population Management.}
We rank all mechanisms in the current population according to their fitness.
The \textit{Select} function randomly selects one or multiple mechanisms $f_l$ in the current population with probability $p_l\propto1/(r_l+N)$, where $r_l$ is its rank and $N$ is the population size.
The \textit{PopulationManager} function selects the $N$ best mechanisms from the current population (comprising parent mechanisms in $M$ and offspring mechanisms in $M'$) to form a population for the next generation.

\subsection{Fitness Evaluation}\label{subsec:fitness}

The fitness evaluation process involves running the resulting mechanisms on a problem setting. 
Following the notations used in Section \ref{sec:pre}, for a given heuristic mechanism $f$ that takes $n$ agent peaks, $\{\tau(u_1),\ldots,\tau(u_n)\}$, as input and outputs the locations of $K$ facilities, $x=\{x_1,\ldots,x_K\}$, we compute its fitness value, $q(f)$, as follows: 
\begin{equation}\label{eq:fitness-function}
q(f)=\frac{1}{R\sum^n_{i=1}\gamma_i}\sum^R_{j=1}\sum^n_{i=1}\gamma_i u_i^{(j)}(f(\cdot))+\theta\left(\max_{i\in N}\text{rgt}_i(f)\right),
\end{equation}
where $R$ is the sample size, $\gamma_i$ is the weight of agent $i$, and $u_i(x)=\min_{k\in\{1,\ldots,K\}}|x_k-\tau(u_i)|$ is the cost of agent $i$, 
defined as the distance from the peak of $i$ to its nearest facility. 
We refer to the first term in Equation \ref{eq:fitness-function} as \emph{weighted social cost} under $f$. 
We use `unweighted' to refer to the special case of $\gamma_i=1$ for all agents, and we may omit `weighted' when discussing the general case.
The (empirical) regret incurred by $f$ to agent $i$, rgt$_i$, is the maximum gain that it can get by misreporting its peaks. Formally,
\begin{align*}
    \text{rgt}_i(f) &= \frac{1}{R}\sum^R_{j=1}\max_{m\in\{1,\ldots,M\}}\left[\max\{0,(\star)\}\right] \\
    (\star) &= u_i(f(\cdot))-u_i(f(\cdot)'),
\end{align*}
where $M$ is the number of misreports from the agents and $f(\cdot)'$ is the returned facility locations had $i$ misreported its peak. 
In other words, the input to $f$ is $\{\tau(u_1),\ldots,\tau(u'_i),\ldots,\tau(u_n)\}$ instead.
We refer to $\max_{i\in N}\text{rgt}_i(f)$ informally as the (empirical) \emph{max regret}, or simply \emph{regret} when the context is clear.
Finally, we define $\theta$ as a function of max regret incurred by $f$:
\begin{equation}\label{eq:fitness-penalty}
\theta(\text{max regret}) = 
\begin{cases}
    1 & \text{if max regret is greater than $\varepsilon$} \\
    0 & \text{otherwise},
\end{cases}
\end{equation}
for some (small) value $0\le\varepsilon\ll1$.
That is, given a heuristic mechanism $f$, we penalize its fitness value (social cost initially) if its associated max regret is greater than some predefined threshold.
Because social cost is bounded within $[0,1]$, a mechanism with a fitness value greater than 1 implies that it is not strategyproof.\footnote{The converse does not hold even when $\varepsilon=0$ for any finite $R$. Hence, `strategyproofness' here is approximate.}
Positive values of $\varepsilon$ relax the strategyproofness constraint. 

\begin{table*}[t]
    \footnotesize
    \centering
    \begin{tabular}{|ccccc|ccllc|}
        \hline
        & $K$ & Per. & Dict. & Cons. & MoulinNet & RegretNet & \multicolumn{1}{c}{LLMMech} & \multicolumn{1}{c}{LLMMech-e} & NonSP \\\hline
        uniform & 1 & 0.20283 & \textbf{0.14948} & 0.24866 & \textbf{0.14948} & 0.14983 (0.00027) & \textbf{0.14948} & \textbf{0.14948} & 0.14948 \\
        uniform & 2 & 0.08121 & 0.06994 & 0.12707 & 0.05382 & 0.06320 (0.00035) & 0.06067 & \textbf{0.05241} & 0.04377 \\
        uniform & 3 & 0.03316 & 0.03354 & 0.08569 & 0.02555 & 0.03912 (0.00088) & 0.03253 & \textbf{0.02085} (0.00173) & 0.01633 \\
        uniform & 4 & 0.01704 & 0.01313 & 0.06388 & 0.01239 & 0.03085 (0.00026) & 0.01293 & \textbf{0.00856} (0.00030) & 0.00473 \\\hline
        normal & 1 & 0.09350 & \textbf{0.07148} & 0.11992 & 0.07150 & 0.07178 (0.00021) & \textbf{0.07148} & \textbf{0.07148} & 0.07148 \\
        normal & 2 & 0.04302 & 0.03360 & 0.07249 & 0.03402 & 0.03503 (0.00038) & \textbf{0.02303} (0.00404) & \textbf{0.02303} (0.00404) & 0.02303 \\
        normal & 3 & 0.01648 & 0.01688 & 0.05570 & 0.01624 & 0.02348 (0.00030) & \textbf{0.01208} & 0.01410 & 0.00865 \\
        normal & 4 & 0.00823 & 0.00694 & 0.03830 & 0.00569 & 0.01902 (0.00015) & \textbf{0.00247} (0.00063) & \textbf{0.00247} (0.00063) & 0.00247 \\\hline
        beta1 & 1 & 0.05422 & \textbf{0.04079} & 0.07173 & \textbf{0.04079} & 0.04094 (0.00009) & \textbf{0.04079} & \textbf{0.04079} & 0.04079 \\
        beta1 & 2 & 0.02396 & 0.01904 & 0.04759 & \textbf{0.01575} & 0.01838 (0.00033) & 0.01844 & 0.01603 & 0.01147 \\
        beta1 & 3 & 0.00866 & 0.00978 & 0.03179 & 0.00655 & 0.00928 (0.00044) & 0.00865 & \textbf{0.00592} & 0.00400 \\
        beta1 & 4 & 0.00305 & 0.00378 & 0.02809 & 0.00188 & 0.00617 (0.00048) & \textbf{0.00110} (0.00017) & \textbf{0.00110} (0.00017) & 0.00110 \\\hline
        beta2 & 1 & 0.06057 & \textbf{0.04128} & 0.07229 & \textbf{0.04128} & 0.04166 (0.00007) & \textbf{0.04128} & \textbf{0.04128} & 0.04128 \\
        beta2 & 2 & 0.01978 & 0.01918 & 0.04756 & 0.01571 & 0.01957 (0.00012) & 0.01305 (0.00079) & \textbf{0.01304} (0.00038) & 0.01164 \\
        beta2 & 3 & 0.00856 & 0.00980 & 0.03201 & 0.00666 & 0.01449 (0.00010) & \textbf{0.00393} (0.00111) & \textbf{0.00393} (0.00111) & 0.00393 \\
        beta2 & 4 & 0.00305 & 0.00407 & 0.02813 & \textbf{0.00189} & 0.01671 (0.00013) & 0.00304 & 0.00267 & 0.00108 \\\hline
    \end{tabular}\\
    \smallskip
    \begin{tabular}{|ccccc|clllc|}
        \hline
        & $K$ & Per. & Dict. & Cons. & MoulinNet & \multicolumn{1}{c}{RegretNet} & \multicolumn{1}{c}{LLMMech} & \multicolumn{1}{c}{LLMMech-e} & NonSP \\\hline
        uniform & 1 & 0.22896 & 0.24250 & 0.25161 & \textbf{0.20888} & 0.21507 (0.00137) & \textbf{0.20888} & \textbf{0.20888} & 0.20888 \\
        uniform & 2 & 0.09926 & 0.09106 & 0.12761 & 0.08175 & 0.08277 (0.00092) & \textbf{0.08146} & \textbf{0.08146} & 0.07285 \\
        uniform & 3 & 0.05879 & 0.05695 & 0.08559 & 0.04455 & 0.04872 (0.00103) & 0.04565 & \textbf{0.04429} & 0.03521 \\
        uniform & 4 & 0.04065 & 0.03866 & 0.06380 & 0.03439 & 0.03736 (0.00073) & 0.03513 & \textbf{0.02794} (0.00096) & 0.02001 \\\hline
        normal & 1 & 0.09537 & 0.10546 & 0.11925 & 0.08795 & 0.09244 (0.00029) & \textbf{0.08791} & \textbf{0.08791} & 0.08791 \\
        normal & 2 & 0.04909 & 0.04124 & 0.08153 & 0.03847 & 0.05340 (0.00060) & 0.03739 (0.00049) & \textbf{0.03673} (0.00063) & 0.03489 \\
        normal & 3 & 0.03041 & 0.02729 & 0.04552 & 0.02485 & 0.02800 (0.00061) & 0.01806 (0.00230) & \textbf{0.01779} (0.00184) & 0.01779 \\
        normal & 4 & 0.02051 & 0.01927 & 0.03512 & 0.01674 & 0.02252 (0.00044) & \textbf{0.01600} & 0.01626 & 0.01012 \\\hline
        beta1 & 1 & 0.06260 & 0.06851 & 0.07157 & 0.05781 & 0.05848 (0.00069) & \textbf{0.05780} & \textbf{0.05780} & 0.05780 \\
        beta1 & 2 & 0.03044 & 0.02677 & 0.04006 & 0.02380 & 0.02378 (0.00085) & \textbf{0.02107} (0.00097) & \textbf{0.02107} (0.00097) & 0.02107 \\
        beta1 & 3 & 0.01720 & 0.01762 & 0.03026 & 0.01474 & 0.01571 (0.00047) & 0.01184 (0.00105) & \textbf{0.01025} (0.00087) & 0.01025 \\
        beta1 & 4 & 0.01035 & 0.01259 & 0.02503 & 0.00799 & 0.01190 (0.00077) & 0.00577 (0.00094) & \textbf{0.00569} (0.00071) & 0.00562 \\\hline
        beta2 & 1 & 0.06073 & 0.06644 & 0.06815 & \textbf{0.05617} & 0.06502 (0.00122) & \textbf{0.05617} & \textbf{0.05617} & 0.05617 \\
        beta2 & 2 & 0.03120 & 0.02538 & 0.03894 & 0.02319 & 0.03777 (0.00070) & \textbf{0.02308} & \textbf{0.02308} & 0.02060 \\
        beta2 & 3 & 0.01663 & 0.01683 & 0.02964 & \textbf{0.01381} & 0.02642 (0.00000) & 0.01438 & 0.01438 & 0.01000 \\
        beta2 & 4 & 0.01016 & 0.01177 & 0.02381 & \textbf{0.00883} & 0.02055 & 0.00929 & 0.00916 & 0.00549 \\\hline
    \end{tabular}
    \caption{Results for $n=5$ (top) and $n=10$ (bottom), weighted social cost. Values in parentheses (if any) represent regret (max across agents) incurred by the mechanism. The best results (beside NonSP) are bolded.}
    \label{tab:res-w-n5+10}
\end{table*}

\subsection{Prompt Evolution}\label{subsec:prompt-evo}

During evolution, the diversity in the population is restricted by the manually designed prompt strategies, which may lead to premature convergence at a local optimum.
To avoid this, we systematically evolve them whenever stagnation occurs.
In particular, when the fitness values of the top-$b$ individuals remain unchanged for $T_p$ consecutive units of evaluation resource (e.g., number of generations), we invoke the prompt evolution process as follows. 
Throughout the evolutionary framework, we record the fitness value of each prompt strategy, defined as the average fitness value of the top-$d$ individuals produced by the respective prompt.
This helps manage the size of the prompt population as more prompt strategies are generated.
When there is stagnation, the prompt population is evolved also via exploration and modification strategies.

\textbf{Exploration:} Given the list of currently used exploration prompts, ask the LLM to design a new exploration prompt that is different from them as much as possible. 

\textbf{Modification:} Given the list of currently used modification prompts, ask the LLM to design a new modification prompt that is different from them as much as possible. 

In other words, new prompt strategies of each type are introduced after triggering prompt evolution.
The \textit{EvolveVariationPrompt} function denotes the generation of offspring prompt strategies by prompting the LLM with parent prompt strategies (of same type) as input.
The exact prompt used is shown in Figure \ref{fig:evol_prompt-design} from Appendix \ref{subapp:prompt-evol}.


\section{Experiments}
\label{sec:exp}


We start this section by describing the datasets used in Section \ref{subsec:setup} and the considered baselines in Section \ref{subsec:baselines}.
Section \ref{subsec:res}  presents our results for demonstrating the efficacy of LLMMech, followed by a closer look at its learned mechanisms in Section \ref{subsec:case-study}.
Finally, we provide an ablation study to further validate our claims in Section \ref{subsec:ablation}.
Appendix \ref{app:res} includes all implementation details and missing results.

\subsection{Experimental Setups}\label{subsec:setup}


In all problem settings, the dataset for both training and testing phases has 1,000 samples with dimensionality of $n$, where each entry represents an agent's peak $\tau(u_i)$. 
Each sample of (true) peaks is associated with 10 misreports, i.e., each agent misreports their peak ten times. 
The peaks are randomly sampled from uniform, normal ($\mu=0.5$, $\sigma=1$), and (two) beta distributions, referred to as `beta1' ($\alpha=1,\beta=9$) and `beta2' ($\alpha=9,\beta=1$), defined on $[0,1]$. 
For clarity, we refer to our LLM framework as LLMMech and the variant with prompt evolution as LLMMech-e.




\begin{figure*}[tbp]
    \centering
    \includegraphics[width=0.99\textwidth]{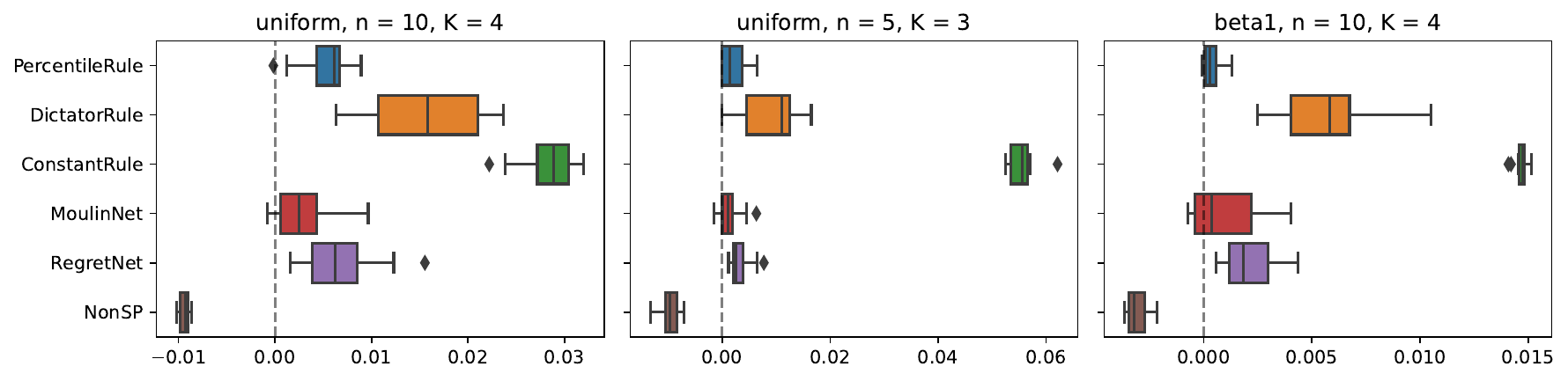}
    \caption{Representative results for arbitrary weights. Each boxplot records net differences in social cost with respect to LLMMech-e.}
    \label{fig:res-aw-short}
\end{figure*}

\begin{table*}[t]
    \footnotesize
    \centering
    \begin{tabular}{|ccccc|ccllc|}
        \hline
        & $K$ & Per. & Dict. & Cons. & MoulinNet & RegretNet@25k & \multicolumn{1}{c}{LLMMech} & \multicolumn{1}{c}{LLMMech-e} & NonSP \\\hline
        uniform & 1 & \textbf{0.20038} & 0.26673 & 0.25242 & 0.20045 & 0.20128 (0.00126) & \textbf{0.20038} & \textbf{0.20038} & 0.20038 \\
        uniform & 2 & 0.08399 & 0.12589 & 0.12708 & 0.10371 & 0.08501 (0.00290) & 0.08398 & \textbf{0.08373} & 0.07098 \\
        uniform & 3 & 0.03328 & 0.06038 & 0.09020 & 0.04027 & 0.03493 (0.00206) & 0.03328 & \textbf{0.03323} (0.00017) & 0.02798 \\
        uniform & 4 & 0.01679 & 0.02324 & 0.06552 & 0.01689 & 0.01999 (0.00156) & \textbf{0.00852} (0.00070) & \textbf{0.00852} (0.00070) & 0.00852 \\\hline
    \end{tabular}
    \caption{Results for $n=5$, unweighted social cost. `RegretNet@25k' denotes RegretNet when trained on 25k samples. Values in parentheses (if any) represent regret (max across agents) incurred by the mechanism. The best results (beside NonSP) are bolded.}
    \label{tab:res-uw-n5-short}
\end{table*}

\subsection{Baseline Methods}\label{subsec:baselines}
We consider the neural networks from \cite{golowich2018deep}, i.e., MoulinNet and RegretNet, as well as the standard mechanisms used in the same work as baselines: the best percentile rule (abbreviated as `Per.') \cite{sui2013analysis}, the best dictatorial rule (`Dict.'), and the best constant rule (`Cons.').
We also present results for the optimal (non-strategyproof) mechanisms (`NonSP').
We refer readers to \cite{golowich2018deep} for further details of these baselines.
We repeat training of both deep learning models with 10 random parameter initializations, 
and choose the network from the run with smallest regret (based only on the training data).
For RegretNet, we fix $\rho$, the hyperparameter controlling the regularization strength for enforcing strategyproofness (in Equation 6 of the paper), to 1, and the batch size to 50.
These values were chosen during our efforts of replicating Table 1 from \cite{golowich2018deep}.

\subsection{Main Results}\label{subsec:res}

\paragraph{Weighted Social Cost with Fixed Weights.}
In this setup, we consider $n=5,10$ and adopt the weight configuration used by \cite{golowich2018deep} where a few agents (one for $n=5$ and two for $n=10$) are assigned a weight of 5 while the remaining agents are assigned a weight of 1.
Table \ref{tab:res-w-n5+10} shows results on the test set for the (best) mechanisms learned by LLMMech, LLMMech-e, and other baselines. 
Either of our LLM methods generally yields the best or second-best social cost with negligible or zero regret.  
We notice whenever $K=1$, the LLM-generated mechanisms always resemble the generalized median rule for one facility location, which is naturally optimal and strategyproof \cite{moulin1980strategy,procaccia2013approximate}. 

\paragraph{Weighted Social Cost with Arbitrary Weights.}
We show our framework can effectively deal with arbitrary weight configurations. In this most general setup, we randomly assign agent weights in $\{1,\ldots,5\}$ for 10 different seeds.
Table \ref{tab:arbi-weights} shows the set of weights used for $n=5$ and $n=10$.
\nate{Note that we run both LLMMech and LLMMech-e with $\varepsilon$ set to 0. In other words, the empirical max regret associated with the reported values is 0.}



Figure \ref{fig:res-aw-short} shows representative results for different combinations of $n$ and $K$ (complete results can be found in Figure \ref{fig:res-aw}), where each boxplot records the net differences in social cost with respect to LLMMech-e (which outperforms vanilla LLMMech) over 10 weight sets. 
We see that even with strict enforcement on strategyproofness, LLMMech-e still generally outperforms other baselines.
Note that given the poor performance of RegretNet as previously observed in Table \ref{tab:res-w-n5+10}, we henceforth augment its training set from 1k to 25k samples (with 10 misreports for each still)\footnote{Although not specified in the main paper, we found the training sample size of RegretNet to be 25k according to the original code from its authors.}.
Additionally, we perform hyperparameter tuning.\footnote{We notice RegretNet with fixed hyperparameters (as specified in Section \ref{subsec:baselines}) severely underperforms for $n=10$ even when trained on 25k samples (see Figure \ref{fig:res-aw-old} in Appendix \ref{app:res}).}
Due to the prohibitively long runtime, we only tuned the batch size $\in\{50,100,200,500\}$ and $\rho\in\{0.1,0.5,1\}$ via Optuna 
\cite{akiba2019optuna}. 
Even through such efforts, RegretNet still trails behind our less-finicky LLMMech-e in all problem settings.

\paragraph{Unweighted Social Cost.}


For completeness, we follow the experiments done in Table 1 of \cite{golowich2018deep} to demonstrate the similarly impressive performance of LLM methods on unweighted settings.
Table \ref{tab:res-uw-n5-short} shows results for $n=5$, $K\in\{1,2,3,4\}$, where LLMMech-e seems to outperform the best percentile rule while MoulinNet and RegretNet could not.
We also extend to $n=10,25$ and evaluate on non-uniform distributions and observe similar patterns (see complete results in Tables \ref{tab:res-uw-n5}, \ref{tab:res-uw-n10}, \ref{tab:res-uw-n25}). 

\begin{table*}[t]
    \footnotesize
    \centering
    \renewcommand{\arraystretch}{1.1}
    \begin{tabular}{|cccccc|cl|c|}
        \hline
        & $K$ & Per. & Dict. & MoulinNet & RegretNet@25k & LLMMech (OOD) & \multicolumn{1}{c|}{LLMMech-e (OOD)} & NonSP \\\hline
        normal & 1 & 0.09350 & \textbf{0.07148} & 0.07150 & 0.07178 (0.00021) & \textbf{0.07148} & \textbf{0.07148} & 0.07148 \\
        normal & 2 & 0.04302 & 0.03360 & 0.03402 & 0.03503 (0.00038) & 0.03097 & \textbf{0.02760} (0.00027) & 0.02303 \\
        normal & 3 & 0.01648 & 0.01688 & 0.01624 & 0.02348 (0.00030) & 0.01628 & \textbf{0.01040} (0.00235) & 0.00865 \\
        normal & 4 & 0.00823 & 0.00694 & 0.00569 & 0.01902 (0.00015) & 0.00699 & \textbf{0.00448} (0.00015) & 0.00247 \\\hline
        beta1 & 1 & 0.05422 & \textbf{0.04079} & \textbf{0.04079} & 0.04094 (0.00009) & \textbf{0.04079} & \textbf{0.04079} & 0.04079 \\
        beta1 & 2 & 0.02396 & 0.01904 & 0.01575 & 0.01838 (0.00033) & 0.01861 & \textbf{0.01508} (0.00026) & 0.01147 \\
        beta1 & 3 & 0.00866 & 0.00978 & 0.00655 & 0.00928 (0.00044) & 0.00850 & \textbf{0.00604} (0.00151) & 0.00400 \\
        beta1 & 4 & 0.00305 & 0.00378 & \textbf{0.00188} & 0.00617 (0.00048) & 0.00390 & 0.00194 (0.00007) & 0.00110 \\\hline
        beta2 & 1 & 0.06057 & \textbf{0.04128} & \textbf{0.04128} & 0.04166 (0.00007) & \textbf{0.04128} & \textbf{0.04128} & 0.04128 \\
        beta2 & 2 & 0.01978 & 0.01918 & 0.01571 & 0.01957 (0.00012) & 0.01683 & \textbf{0.01324} & 0.01164 \\
        beta2 & 3 & 0.00856 & 0.00980 & 0.00666 & 0.01449 (0.00010) & 0.00792 & \textbf{0.00555} & 0.00393 \\
        beta2 & 4 & 0.00305 & 0.00407 & \textbf{0.00189} & 0.01671 (0.00013) & 0.00392 & 0.00203 (0.00007) & 0.00108 \\\hline
    \end{tabular}\\
    \smallskip
    \begin{tabular}{|cccccl|cl|c|}
        \hline
        & $K$ & Per. & Dict. & MoulinNet & \multicolumn{1}{c|}{RegretNet@25k} & LLMMech (OOD) & \multicolumn{1}{c|}{LLMMech-e (OOD)} & NonSP \\\hline
        normal & 1 & 0.09537 & 0.10546 & 0.08795 & 0.09244 (0.00029) & \textbf{0.08791} & \textbf{0.08791} & 0.08791 \\
        normal & 2 & 0.04909 & 0.04124 & 0.03847 & 0.05340 (0.00060) & \textbf{0.03802} & \textbf{0.03802} & 0.03489 \\
        normal & 3 & 0.03041 & 0.02729 & 0.02485 & 0.02800 (0.00061) & 0.02310 & \textbf{0.02228} & 0.01779 \\
        normal & 4 & 0.02051 & 0.01927 & 0.01674 & 0.02252 (0.00044) & 0.01822 & \textbf{0.01471} (0.00053) & 0.01012 \\\hline
        beta1 & 1 & 0.06260 & 0.06851 & 0.05781 & 0.05848 (0.00069) & \textbf{0.05780} & \textbf{0.05780} & 0.05780 \\
        beta1 & 2 & 0.03044 & 0.02677 & 0.02380 & 0.02378 (0.00085) & 0.02405 & \textbf{0.02361} (0.00070) & 0.02107 \\
        beta1 & 3 & 0.01720 & 0.01762 & \textbf{0.01474} & 0.01571 (0.00047) & 0.01485 & 0.01475 & 0.01025 \\
        beta1 & 4 & 0.01035 & 0.01259 & \textbf{0.00799} & 0.01190 (0.00077) & 0.01227 & 0.00949 & 0.00562 \\\hline
        beta2 & 1 & 0.06073 & 0.06644 & \textbf{0.05617} & 0.06502 (0.00122) & \textbf{0.05617} & \textbf{0.05617} & 0.05617 \\
        beta2 & 2 & 0.03120 & 0.02538 & 0.02319 & 0.03777 (0.00070) & 0.02342 & \textbf{0.02289} (0.00041) & 0.02060 \\
        beta2 & 3 & 0.01663 & 0.01683 & 0.01381 & 0.02642 (0.00000) & \textbf{0.01294} & 0.01438 & 0.01000 \\
        beta2 & 4 & 0.01016 & 0.01177 & 0.00883 & 0.02055 & 0.01179 & \textbf{0.00799} (0.00026) & 0.00549 \\\hline
    \end{tabular}
    \caption{Out-of-distribution (OOD) results for $n=5$ (top) and $n=10$ (bottom), weighted social cost. Baselines were trained with in-distribution (uniformly distributed) samples. Values in parentheses (if any) represent regret (max across agents) incurred by the mechanism. The best results (beside NonSP) are bolded.}
    \label{tab:res-w-n5+10-ood}
\end{table*}

\subsection{A Case Study on Learned Mechanisms}\label{subsec:case-study}

Our results show that LLMMech and especially LLMMech-e are capable of generating novel, empirically strategyproof mechanisms that outperform existing methods.
Figure \ref{fig:learned-mech} exemplifies such mechanism for the setting of $n=5,K=2$ with uniform peak distribution. (Its fitness value is omitted and the associated Python code is available in Appendix \ref{app:designed-mechs}.) 
Let $\tau(u_1) \le \tau(u_2) \le \tau(u_3) \le \tau(u_4) \le \tau(u_5)$ be the agent peaks.
Essentially, the mechanism first locates the first facility at the weight median of left 50\% agents ($i=1,2$), then locates the second facility at the weight median of right 50\% agents ($i=3,4,5$).
In this particular problem setting, 
the mechanism is strategyproof because it always outputs the facility at $x_1=\tau(u_1)$ and $x_2=\tau(u_4)$. 
However, it is not strategyproof in general: Consider an instance with agent weights $[5,5,5,1,1,1]$ and peaks $[0,1/13,10/13,11/13,12/13,1]$.  
The mechanism outputs $x=[1/13,12/13]$ when all agents report their peaks truthfully. If the third agent at $10/13$ misreports its peak to $11/13<\tau(u'_3)<12/13$, the mechanism will instead output $x'=[1/13,\tau(u'_3)]$, which benefits the agent. 

Our case study demonstrates the importance of interpretability that is unattainable for deep learning models.
Even when they yield zero empirical regret, we would not be able to verify their strategyproofness on unseen problem instances.

\begin{figure}[h]
    \footnotesize
    \centering
    \begin{tcolorbox}[width=0.49\textwidth, colback=blue!5]
        \textbf{Mechanism Description} (Generated via LLMMech-e)

        \hphantom\\

        A hybrid mechanism that combines elements of clustering and optimization to identify two facility locations by first clustering the samples into groups based on proximity and then selecting the best locations within these groups to minimize utility values.
        \end{tcolorbox}
    \caption{High-level description of a learned mechanism that yields the lowest cost compared to other baselines (2nd row of Table \ref{tab:res-w-n5+10}).}
    \label{fig:learned-mech}
\end{figure}

\begin{figure}[h]
    \footnotesize
    \centering
    \begin{tcolorbox}[width=0.49\textwidth, colback=blue!5]
        \textbf{(New) Prompt Strategy}

        \hphantom\\

        \textcolor{mauve}{Generate a hybrid mechanism that combines elements of clustering and optimization techniques to identify two facility locations based on the provided location samples, while balancing both local and global utility values.}
        \end{tcolorbox}
    \caption{LLM-generated exploration prompt strategy that invokes design of the novel mechanism described in Figure \ref{fig:learned-mech}.}
    \label{fig:learned-prompt}
\end{figure}

\begin{table*}[t]
    \footnotesize
    \centering
    \renewcommand{\arraystretch}{1.15}
    \begin{tabular}{|ccccl|ll|c|}
        \hline
        & $K$ & Per. & MoulinNet & \multicolumn{1}{c|}{RegretNet@25k} & \multicolumn{1}{c}{LLMMech ($n=5$)} & \multicolumn{1}{c|}{LLMMech-e ($n=5$)} & NonSP \\\hline
        uniform & 1 & \textbf{0.22759} & 0.22764 & 0.24228 (0.00264) & \textbf{0.22759} & \textbf{0.22759} & 0.22759 \\
        uniform & 2 & \textbf{0.10004} & 0.10009 & 0.11796 (0.00310) & \textbf{0.10004} & \textbf{0.10004} & 0.09458 \\
        uniform & 3 & \textbf{0.05922} & 0.05926 & 0.07993 (0.00201) & 0.07261 & 0.07257 (0.00102) & 0.05196 \\
        uniform & 4 & 0.04096 & 0.04899 & 0.06046 (0.00221) & 0.04165 (0.00640) & \textbf{0.03421} (0.00439) & 0.03173 \\\hline
        normal & 1 & \textbf{0.09585} & \textbf{0.09585} & 0.10504 (0.00102) & \textbf{0.09585} & \textbf{0.09585} & 0.09585 \\
        normal & 2 & \textbf{0.04955} & 0.04981 & 0.06361 (0.00236) & 0.05320 & \textbf{0.04955} & 0.04629 \\
        normal & 3 & 0.03067 & 0.03091 & 0.04786 (0.00187) & 0.03589 & \textbf{0.02666} (0.00242) & 0.02651 \\
        normal & 4 & \textbf{0.02042} & 0.02136 & 0.03611 (0.00119) & 0.02091 (0.00352) & 0.02091 (0.00352) & 0.01594 \\\hline
        beta1 & 1 & \textbf{0.06310} & 0.06311 & 0.06436 (0.00192) & \textbf{0.06310} & \textbf{0.06310} & 0.06310 \\
        beta1 & 2 & \textbf{0.03022} & 0.03056 & 0.03483 (0.00292) & 0.03078 & 0.03078 & 0.02794 \\
        beta1 & 3 & 0.01703 & 0.01726 & 0.02251 (0.00252) & \textbf{0.01493} (0.00133) & \textbf{0.01493} (0.00133) & 0.01493 \\
        beta1 & 4 & 0.01029 & 0.01047 & 0.01638 (0.00154) & 0.01140 (0.00172) & \textbf{0.00941} (0.00110) & 0.00866 \\\hline
        beta2 & 1 & \textbf{0.06092} & \textbf{0.06092} & 0.06411 (0.00246) & \textbf{0.06092} & \textbf{0.06092} & 0.06092 \\
        beta2 & 2 & \textbf{0.02934} & 0.03243 & 0.03764 (0.00140) & 0.03296 & 0.02989 & 0.02720 \\
        beta2 & 3 & 0.01668 & 0.01670 & 0.02641 (0.00042) & \textbf{0.01461} (0.00082) & \textbf{0.01461} (0.00082) & 0.01461 \\
        beta2 & 4 & \textbf{0.01031} & 0.01239 & 0.02070 & 0.01575 & 0.01175 (0.00047) & 0.00847 \\\hline
    \end{tabular}
    \caption{Results for generalizability of LLM-generated mechanisms from $n=5$ to $n=10$, unweighted social cost. Baselines were trained directly on $n=10$ dataset. Values in parentheses (if any) represent regret (max across agents) incurred by the mechanism. The best results (beside NonSP) are bolded.}
    \label{tab:res-uw-n10-generalizability}
\end{table*}

\paragraph{An Example of Learned Prompts.}
We next scrutinize the prompt strategy behind our previously considered mechanism.
As expected, the prompt strategy, illustrated in Figure \ref{fig:learned-prompt}, was newly introduced during the prompt evolution process.
In addition to the overall better performance of LLMMech-e with respect to vanilla LLMMech (Table \ref{tab:res-w-n5+10}), this observation further supports the need of an automatic evolution of variation prompts in order to diversify the mechanism population and in turn help the search process escape from local optima.


\subsection{Ablation Study}\label{subsec:ablation}

\paragraph{Out-of-Distribution Performance.}

We investigate the out-of-distribution (OOD) performance of LLM methods.
In particular, given the learned mechanisms from LLMMech and LLMMech-e that are trained on uniformly distributed peaks, we test them on instances with peaks generated from other distributions.
Table \ref{tab:res-w-n5+10-ood} shows results for $n=5$ and 10 with fixed weights. 
In general, LLM methods achieve similar or better performance with respect to all baselines evaluated on in-distribution peaks.
Please refer to Tables \ref{tab:res-w-n5-ood} and \ref{tab:res-w-n10-ood} for comparison with baselines when they are also evaluated on OOD peaks. 


\paragraph{Generalizability to Larger Instances.}

We also examine the generalizability of LLM-generated mechanisms to larger instances with more agents.
Note that this experiment makes senses for unweighted settings only.
Table \ref{tab:res-uw-n10-generalizability} shows results for $n=10$, where LLM methods are trained on $n=5$ instances (see also Table \ref{tab:res-uw-n25-generalizability} for $n=25$).
We observe impressive generalization performance from both LLMMech and LLMMech-e, with social costs generally close to or in some settings (e.g., $n=10$ and $K=4$ for uniform and beta1 distributions) even better than other methods\footnote{Handcrafted rules and neural networks are restricted to fixed input size and hence only applicable to similar-sized instances.}.
This overall generalizability allows the learned mechanisms to quickly adapt to larger problems with minimal computational overheads.
In contrast, the neural networks need to be retrained from scratch and are highly inefficient for large input size (i.e., $n$), whereas the manually designed rules also require recomputation, which are as costly given their brute-force nature. 




\section{Conclusion}
\label{sec:conclusion}

In this paper, we propose LLMMech, the first LLM-based evolutionary framework for the automatic design of multi-facility location mechanisms. 
Moreover, we demonstrate through our various experiments that, with minimal domain knowledge and hyperparameter tuning, LLMMech can generate empirically strategyproof 
and nearly optimal mechanisms that can generalize to out-of-distribution agent preferences and to larger instances with more agents. 
By this means, LLMMech can significantly reduce the manual effort involved in traditional mechanism design without compromising interpretability, which is crucial when transparency of the resulting mechanisms is required during social planning. 
\nate{Future work could extend our framework to design mechanisms for more general facility location problems, e.g., considering fairness/envy objectives, higher dimensional space, facilities with constraints, and other types of agent preferences.}


\newpage

\section*{Impact Statement}
This paper presents work whose goal is to advance the field of Machine Learning. There are many potential societal consequences of our work, none of which we feel must be specifically highlighted here.

\newpage


\bibliography{main}

\begin{thebibliography}{23}
\providecommand{\natexlab}[1]{#1}
\providecommand{\url}[1]{\texttt{#1}}
\expandafter\ifx\csname urlstyle\endcsname\relax
  \providecommand{\doi}[1]{doi: #1}\else
  \providecommand{\doi}{doi: \begingroup \urlstyle{rm}\Url}\fi

\bibitem[Ahmadi-Javid et~al.(2017)Ahmadi-Javid, Seyedi, and Syam]{Ahmadi-Javid:2017aa}
Ahmadi-Javid, A., Seyedi, P., and Syam, S.~S.
\newblock A survey of healthcare facility location.
\newblock \emph{Computers \& Operations Research}, 79:\penalty0 223--263, 2017.

\bibitem[Ahn et~al.(2024)Ahn, Verma, Lou, Liu, Zhang, and Yin]{ahn2024large}
Ahn, J., Verma, R., Lou, R., Liu, D., Zhang, R., and Yin, W.
\newblock Large language models for mathematical reasoning: Progresses and challenges.
\newblock \emph{arXiv preprint arXiv:2402.00157}, 2024.

\bibitem[Akiba et~al.(2019)Akiba, Sano, Yanase, Ohta, and Koyama]{akiba2019optuna}
Akiba, T., Sano, S., Yanase, T., Ohta, T., and Koyama, M.
\newblock {O}ptuna: A next-generation hyperparameter optimization framework.
\newblock In \emph{The 25th ACM SIGKDD International Conference on Knowledge Discovery \& Data Mining}, pp.\  2623--2631, 2019.

\bibitem[Black(1948)]{Black:1948aa}
Black, D.
\newblock On the rationale of group decision-making.
\newblock \emph{Journal of Political Economy}, 56\penalty0 (1):\penalty0 23--34, 1948.

\bibitem[Brown et~al.(2020)Brown, Mann, Ryder, Subbiah, Kaplan, Dhariwal, Neelakantan, Shyam, Sastry, Askell, et~al.]{brown2020language}
Brown, T., Mann, B., Ryder, N., Subbiah, M., Kaplan, J.~D., Dhariwal, P., Neelakantan, A., Shyam, P., Sastry, G., Askell, A., et~al.
\newblock Language models are few-shot learners.
\newblock \emph{Advances in neural information processing systems}, 33:\penalty0 1877--1901, 2020.

\bibitem[Chan et~al.(2021)Chan, Filos-Ratsikas, Li, Li, and Wang]{Chan:2021aa}
Chan, H., Filos-Ratsikas, A., Li, B., Li, M., and Wang, C.
\newblock Mechanism design for facility location problems: A survey.
\newblock In Zhou, Z.-H. (ed.), \emph{Proceedings of the Thirtieth International Joint Conference on Artificial Intelligence, {IJCAI-21}}, pp.\  4356--4365, 2021.

\bibitem[Escoffier et~al.(2011)Escoffier, Gourves, Kim~Thang, Pascual, and Spanjaard]{escoffier2011strategy}
Escoffier, B., Gourves, L., Kim~Thang, N., Pascual, F., and Spanjaard, O.
\newblock Strategy-proof mechanisms for facility location games with many facilities.
\newblock In \emph{Algorithmic Decision Theory: Second International Conference, ADT 2011, Piscataway, NJ, USA, October 26-28, 2011. Proceedings 2}, pp.\  67--81. Springer, 2011.

\bibitem[Farahani \& Hekmatfar(2009)Farahani and Hekmatfar]{Farahani:2009aa}
Farahani, R.~Z. and Hekmatfar, M.
\newblock \emph{Facility location: concepts, models, algorithms and case studies}.
\newblock Springer Science \& Business Media, 2009.

\bibitem[Fotakis \& Tzamos(2014)Fotakis and Tzamos]{fotakis2014power}
Fotakis, D. and Tzamos, C.
\newblock On the power of deterministic mechanisms for facility location games.
\newblock \emph{ACM Transactions on Economics and Computation (TEAC)}, 2\penalty0 (4):\penalty0 1--37, 2014.

\bibitem[Golowich et~al.(2018)Golowich, Narasimhan, and Parkes]{golowich2018deep}
Golowich, N., Narasimhan, H., and Parkes, D.~C.
\newblock Deep learning for multi-facility location mechanism design.
\newblock In \emph{IJCAI}, pp.\  261--267, 2018.

\bibitem[Hu et~al.(2024)Hu, Lu, and Clune]{hu2024automated}
Hu, S., Lu, C., and Clune, J.
\newblock Automated design of agentic systems.
\newblock \emph{arXiv preprint arXiv:2408.08435}, 2024.

\bibitem[Jiang et~al.(2024)Jiang, Wang, Shen, Kim, and Kim]{jiang2024survey}
Jiang, J., Wang, F., Shen, J., Kim, S., and Kim, S.
\newblock A survey on large language models for code generation.
\newblock \emph{arXiv preprint arXiv:2406.00515}, 2024.

\bibitem[Liu et~al.(2023)Liu, Tong, Yuan, and Zhang]{liu2023algorithm}
Liu, F., Tong, X., Yuan, M., and Zhang, Q.
\newblock Algorithm evolution using large language model.
\newblock \emph{arXiv preprint arXiv:2311.15249}, 2023.

\bibitem[Liu et~al.(2024{\natexlab{a}})Liu, Xialiang, Yuan, Lin, Luo, Wang, Lu, and Zhang]{liu2024evolution}
Liu, F., Xialiang, T., Yuan, M., Lin, X., Luo, F., Wang, Z., Lu, Z., and Zhang, Q.
\newblock Evolution of heuristics: Towards efficient automatic algorithm design using large language model.
\newblock In \emph{Forty-first International Conference on Machine Learning}, 2024{\natexlab{a}}.

\bibitem[Liu et~al.(2024{\natexlab{b}})Liu, Yao, Guo, Yang, Lin, Tong, Yuan, Lu, Wang, and Zhang]{liu2024systematic}
Liu, F., Yao, Y., Guo, P., Yang, Z., Lin, X., Tong, X., Yuan, M., Lu, Z., Wang, Z., and Zhang, Q.
\newblock A systematic survey on large language models for algorithm design.
\newblock \emph{arXiv preprint arXiv:2410.14716}, 2024{\natexlab{b}}.

\bibitem[Liu et~al.(2024{\natexlab{c}})Liu, Zhang, Xie, Sun, Li, Lin, Wang, Lu, and Zhang]{liu2024llm4ad}
Liu, F., Zhang, R., Xie, Z., Sun, R., Li, K., Lin, X., Wang, Z., Lu, Z., and Zhang, Q.
\newblock Llm4ad: A platform for algorithm design with large language model.
\newblock \emph{arXiv preprint arXiv:2412.17287}, 2024{\natexlab{c}}.

\bibitem[Lu et~al.(2010)Lu, Sun, Wang, and Zhu]{lu2010asymptotically}
Lu, P., Sun, X., Wang, Y., and Zhu, Z.~A.
\newblock Asymptotically optimal strategy-proof mechanisms for two-facility games.
\newblock In \emph{Proceedings of the 11th ACM conference on Electronic commerce}, pp.\  315--324, 2010.

\bibitem[Moulin(1980)]{moulin1980strategy}
Moulin, H.
\newblock On strategy-proofness and single peakedness.
\newblock \emph{Public Choice}, 35\penalty0 (4):\penalty0 437--455, 1980.

\bibitem[Procaccia \& Tennenholtz(2013)Procaccia and Tennenholtz]{procaccia2013approximate}
Procaccia, A.~D. and Tennenholtz, M.
\newblock Approximate mechanism design without money.
\newblock \emph{ACM Transactions on Economics and Computation (TEAC)}, 1\penalty0 (4):\penalty0 1--26, 2013.

\bibitem[Romera-Paredes et~al.(2024)Romera-Paredes, Barekatain, Novikov, Balog, Kumar, Dupont, Ruiz, Ellenberg, Wang, Fawzi, et~al.]{romera2024mathematical}
Romera-Paredes, B., Barekatain, M., Novikov, A., Balog, M., Kumar, M.~P., Dupont, E., Ruiz, F.~J., Ellenberg, J.~S., Wang, P., Fawzi, O., et~al.
\newblock Mathematical discoveries from program search with large language models.
\newblock \emph{Nature}, 625\penalty0 (7995):\penalty0 468--475, 2024.

\bibitem[Sui et~al.(2013)Sui, Boutilier, and Sandholm]{sui2013analysis}
Sui, X., Boutilier, C., and Sandholm, T.
\newblock Analysis and optimization of multi-dimensional percentile mechanisms.
\newblock In \emph{IJCAI}, pp.\  367--374. Citeseer, 2013.

\bibitem[Wang et~al.(2023)Wang, Fu, Du, Gao, Huang, Liu, Chandak, Liu, Van~Katwyk, Deac, et~al.]{wang2023scientific}
Wang, H., Fu, T., Du, Y., Gao, W., Huang, K., Liu, Z., Chandak, P., Liu, S., Van~Katwyk, P., Deac, A., et~al.
\newblock Scientific discovery in the age of artificial intelligence.
\newblock \emph{Nature}, 620\penalty0 (7972):\penalty0 47--60, 2023.

\bibitem[Zhang et~al.(2024)Zhang, Liu, Lin, Wang, Lu, and Zhang]{zhang2024understanding}
Zhang, R., Liu, F., Lin, X., Wang, Z., Lu, Z., and Zhang, Q.
\newblock Understanding the importance of evolutionary search in automated heuristic design with large language models.
\newblock In \emph{International Conference on Parallel Problem Solving from Nature}, pp.\  185--202. Springer, 2024.

\end{thebibliography}
\bibliographystyle{icml2025}

\newpage
\appendix
\onecolumn

\renewcommand{\thefigure}{S\arabic{figure}}
\renewcommand{\thetable}{S\arabic{table}}
\renewcommand{\theequation}{S\arabic{equation}}

\section{Related Work}

In this appendix, we discuss related work focusing on mechanism design for facility location and LLMs for solving complex problems. 

\paragraph{Mechanism Design for Facility Location.}
\citet{moulin1980strategy} first developed a complete characterization of all strategyproof
mechanisms for locating a facility where agents have single-peaked preferences on a real line. \citet{procaccia2013approximate} first considered strategyproof, approximately optimal mechanisms in facility location on the real line. For locating a single facility, they showed that locating the facility at the median agent location is the optimal mechanism for minimizing the total distance of all agents (i.e., the considered social cost with unweighted agents). 
For locating two facilities with $n$ agents, \citet{procaccia2013approximate} gave a lower-bound of $3/2-O(1/n)$ and an upper-bound of $n-2$ for deterministic strategyproof mechanisms. 
The lower bound was later improved to asymptotically and matching tight bounds of $\Omega(n)$ and $n-2$ \citep{lu2010asymptotically,fotakis2014power}. 
Beyond two facilities, all existing strategyproof mechanisms have an upper-bound of $\Omega(n)$ \cite{fotakis2014power,escoffier2011strategy}. 
\citet{golowich2018deep} utilized neural networks to learn and design mechanisms for locating multiple facilities on the real line and optimizing the considered social cost with unweighted agents. 
We refer readers to a recent survey for more details \cite{Chan:2021aa}. 

\paragraph{Large Language Models.}

LLMs have been applied in diverse tasks, including mathematical reasoning~\citep{ahn2024large}, code generation~\citep{jiang2024survey}, scientific discovery~\citep{wang2023scientific}, and algorithm design~\cite{liu2024systematic,liu2024llm4ad}. EoH~\cite{liu2023algorithm,liu2024evolution} proposed to adopt LLMs to automatically generate, combine and improve heuristics in an evolutionary search. FunSearch~\cite{romera2024mathematical} integrated LLMs into a multi-island iterative search framework for function design. \citep{hu2024automated} proposed ADAS, or Automated Design of Agentic Systems, that aims to automatically generate powerful agentic system designs by using meta agents to program new agents. The significance of integrating LLMs into iterative search frameworks, particularly for challenging design tasks requiring environmental evaluations, has been emphasized in~\cite{zhang2024understanding}. Despite these advancements~\cite{liu2024systematic}, the application of LLMs to multi-facility location mechanism design remains unexplored.


\section{Prompts}

\subsection{Variation Prompts}\label{subapp:var-prompt}
\begin{figure*}[h]
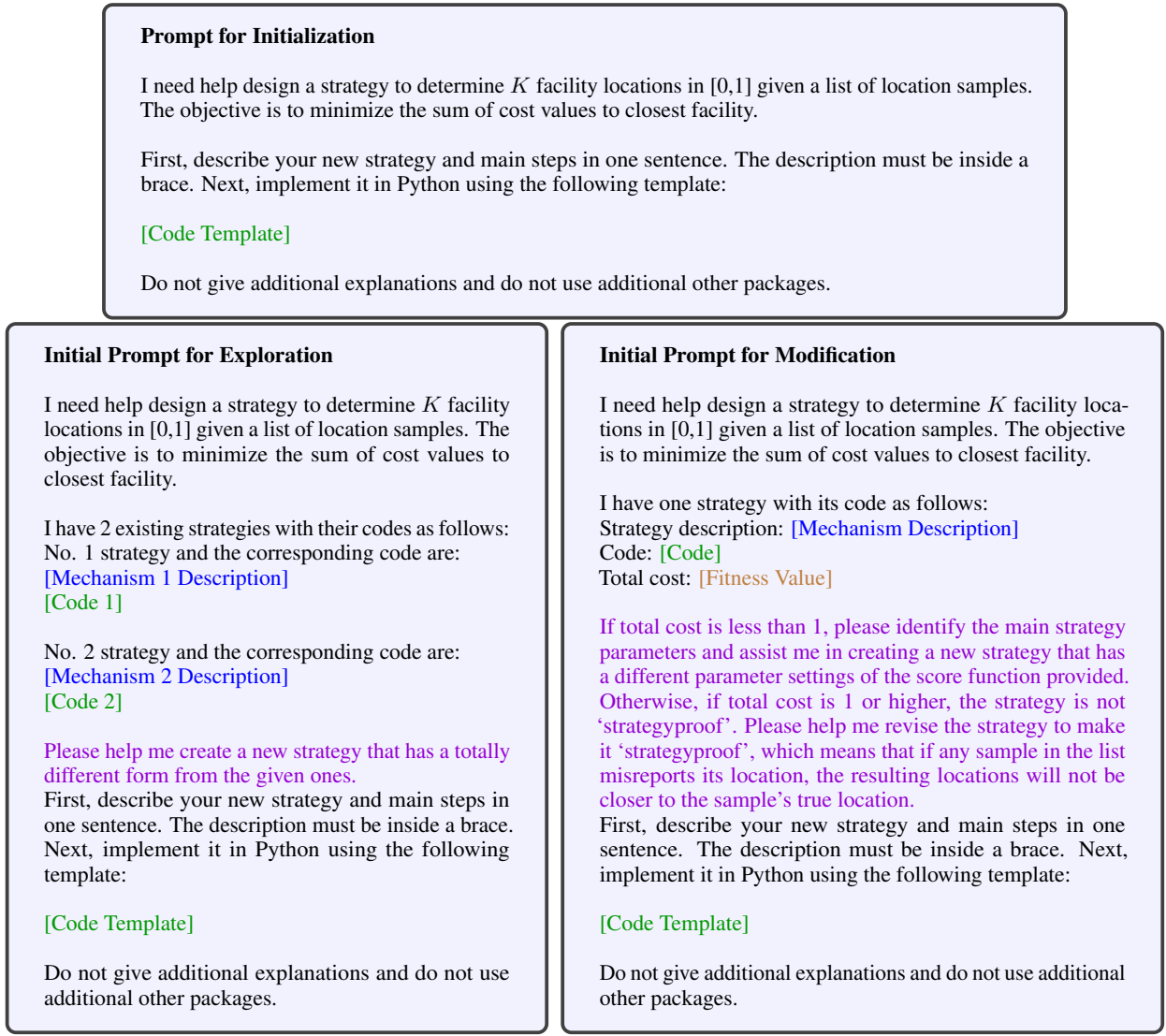

    \centering
    \begin{subfigure}[h]{0.8\textwidth}
        \centering
        \footnotesize
        \begin{tcolorbox}[width=\textwidth, colback=blue!5]
        \textbf{Prompt for Initialization}

        \hphantom\\

        I need help design a strategy to determine $K$ facility locations in [0,1] given a list of location samples. The objective is to minimize the sum of cost values to closest facility.

        \hphantom\\

        First, describe your new strategy and main steps in one sentence. The description must be inside a brace. Next, implement it in Python using the following template:

        \hphantom\\

        \textcolor{dkgreen}{[Code Template]}

        \hphantom\\
        
        Do not give additional explanations and do not use additional other packages.
        \end{tcolorbox}
    \end{subfigure}\\
    \begin{subfigure}[h]{0.45\textwidth}
        \centering
        \footnotesize
        \begin{tcolorbox}[width=\textwidth, colback=blue!5]
        \textbf{Initial Prompt for Exploration}

        \hphantom\\

        I need help design a strategy to determine $K$ facility locations in [0,1] given a list of location samples. The objective is to minimize the sum of cost values to closest facility.

        \hphantom\\

        I have 2 existing strategies with their codes as follows:\\
        No. 1 strategy and the corresponding code are:\\
        \textcolor{blue}{[Mechanism 1 Description]}\\
        \textcolor{dkgreen}{[Code 1]}

        \hphantom\\

        No. 2 strategy and the corresponding code are:\\
        \textcolor{blue}{[Mechanism 2 Description]}\\
        \textcolor{dkgreen}{[Code 2]}

        \hphantom\\

        \textcolor{mauve}{Please help me create a new strategy that has a totally different form from the given ones.}\\
        First, describe your new strategy and main steps in one sentence. The description must be inside a brace. Next, implement it in Python using the following template:

        \hphantom\\

        \textcolor{dkgreen}{[Code Template]}

        \hphantom\\
        
        Do not give additional explanations and do not use additional other packages.
        \end{tcolorbox}
    \end{subfigure}%
    ~
    \begin{subfigure}[h]{0.5\textwidth}
        \centering
        \footnotesize
        \begin{tcolorbox}[width=\textwidth, colback=blue!5]
        \textbf{Initial Prompt for Modification}

        \hphantom\\

        I need help design a strategy to determine $K$ facility locations in [0,1] given a list of location samples. The objective is to minimize the sum of cost values to closest facility.

        \hphantom\\

        I have one strategy with its code as follows:\\
        Strategy description:  \textcolor{blue}{[Mechanism Description]}\\
        Code: \textcolor{dkgreen}{[Code]}\\
        Total cost: \textcolor{brown}{[Fitness Value]}

        \hphantom\\

        \textcolor{mauve}{If total cost is less than 1, please identify the main strategy parameters and assist me in creating a new strategy that has a different parameter settings of the score function provided. Otherwise, if total cost is 1 or higher, the strategy is not `strategyproof'. Please help me revise the strategy to make it `strategyproof', which means that if any sample in the list misreports its location, the resulting locations will not be closer to the sample's true location.}\\
        First, describe your new strategy and main steps in one sentence. The description must be inside a brace. Next, implement it in Python using the following template:

        \hphantom\\

        \textcolor{dkgreen}{[Code Template]}

        \hphantom\\
        
        Do not give additional explanations and do not use additional other packages.
        \end{tcolorbox}
    \end{subfigure}
    \caption{Prompts used for initialization, exploration, and modification. The prompt strategies are marked in purple. See Appendix \ref{app:code-template} for the designed ``Code Template''.}
    \label{fig:init_prompt-design}
\end{figure*}

Figure \ref{fig:init_prompt-design} illustrates the exact initialization prompt and the initial prompts for exploration and modification.

\clearpage
\subsection{Prompt Evolution}\label{subapp:prompt-evol}
Figure \ref{fig:evol_prompt-design} illustrates the exact prompt used for evolving prompt strategies.

\begin{figure*}[h]
    \centering
    \begin{subfigure}[h]{0.9\textwidth}
        \centering
        \footnotesize
        \begin{tcolorbox}[width=\textwidth, colback=blue!5]
        \textbf{Prompt for Evolving Prompt Strategies}

        \hphantom\\

        I need help design a strategy to determine $K$ facility locations in [0,1] given a list of location samples. The objective is to minimize the sum of cost values to closest facility.

        \hphantom\\

        I want to leverage the capabilities of LLMs to generate heuristic algorithms that can efficiently tackle this problem. I have already developed a set of initial prompts and observed the corresponding outputs. However, to improve the effectiveness of these algorithms, we need your assistance in carefully analyzing the existing prompts and their results. Based on this analysis, we ask you to generate new prompts that will help us achieve better results in solving the problem.

        \hphantom\\

        I have X existing prompts with average score (the lower the better) as follows:\\
        No. 1 prompt:\\
        Content: \textcolor{mauve}{[Prompt Strategy]}\\
        Score: \textcolor{orange}{[Prompt Fitness Value]}\\
        ...\\
        No. X prompt:\\
        Content: \textcolor{mauve}{[Prompt Strategy]}\\
        Score: \textcolor{orange}{[Prompt Fitness Value]}

        \hphantom\\

        Note that we categorize prompts into two groups: Exploration and Modification. Those I just showed are \textcolor{purple}{[Type]} prompts, which ask LLMs to generate new strategies that are as different as possible from the input strategies.

        \hphantom\\

        Please help me create a new \textcolor{purple}{[Type]} prompt that has a totally different form from the given ones but can be motivated from them.\\
        Describe your new prompt and main steps in one sentence. The description must be inside a brace. Do not give additional explanations.
        \end{tcolorbox}
    \end{subfigure}%
    \caption{Prompts used for evolving prompt strategies, where \textcolor{purple}{[Type]} is `Exploration' or `Modification' and X $=|\bm{P}|\le N_p$.}
    \label{fig:evol_prompt-design}
\end{figure*}

\clearpage
\section{Code Template}\label{app:code-template}

\begin{lstlisting}[caption={Code template for unweighted settings.},captionpos=b]
def get_locations(samples):
    '''
    Determines the optimal locations from a given list of location samples.

    Args:
    samples (list): A one-dimensional list containing the location samples.

    Returns:
    list: A one-dimensional list of the optimal locations, containing n_locations elements in [0,1].
    '''

    # Placeholder (replace with your actual implementation)
    locations = ...

    return locations
\end{lstlisting}

\begin{lstlisting}[caption={Code template for weighted settings, shown here for the case of $n=5$, fixed weights.},captionpos=b]
def get_locations(samples):
    '''
    Determines the optimal locations from a given list of location samples.

    Args:
    samples (list): A one-dimensional list containing the location samples.
    weights (list): A list of fixed weights assigned to the samples: [5,1,1,1,1] 

    Returns:
    list: A one-dimensional list of the optimal locations, containing n_locations elements in [0,1].
    '''

    # Placeholder (replace with your actual implementation)
    locations = ...

    return locations
\end{lstlisting}

\clearpage
\section{Designed Mechanisms}\label{app:designed-mechs}

\begin{lstlisting}[caption={Learnerd mechanism for $n=5,K=2$, weighted agents with uniformly distributed peaks.},captionpos=b]
def get_locations(samples):
    '''
    Determines the optimal locations from a given list of location samples.

    Args:
    samples (list): A one-dimensional list containing the location samples.
    weights (list): A list of fixed weights assigned to the samples: [5,1,1,1,1] 

    Returns:
    list: A one-dimensional list of the optimal locations, containing n_locations elements in [0,1].
    '''
    weights = [5] + [1] * (len(samples) - 1)
    weighted_samples = list(zip(samples, weights))
    
    # Step 1: Cluster the samples into two groups based on proximity
    weighted_samples.sort()  # Sort samples based on location
    mid_index = len(weighted_samples) // 2

    group1 = weighted_samples[:mid_index]
    group2 = weighted_samples[mid_index:]

    # Step 2: Calculate the weighted median for both groups
    def weighted_median(group):
        total_weight = sum(weight for _, weight in group)
        cum_weight = 0
        for sample, weight in group:
            cum_weight += weight
            if cum_weight >= total_weight / 2:
                return sample
    
    facility1 = weighted_median(group1)
    facility2 = weighted_median(group2)

    locations = [facility1, facility2]
    
    return locations
\end{lstlisting}


\section{Additional Experiments}\label{app:res}

All experiments were conducted under Ubuntu 20.04 on a Linux virtual machine equipped with NVIDIA GeForce RTX 3050 Ti GPU and 12th Gen Intel(R) Core(TM) i7-12700H CPU @2.3GHz.
We chose GPT-4o mini as our pretrained LLM.
The code for our implementation in Python 3.10 is \nate{uploaded as supplementary material}. 

\subsection{Implementation Details}\label{subapp:imp}
We run three trials of LLMMech for each problem configuration, and three trials of LLMMech-e for each LLMMech trial (i.e., evolving directly from the initial population of LLMMech).
In both versions, the temperature is fixed at 1.
The evaluation budget $T_e$ is defined in terms of the maximum number of generations, for which we set to 20 for all problems. 
The population size is 16, and the maximum running time for each instance is 60 seconds (except when $n=25$, for which we increase to 180). 
Unless otherwise stated, we assign the threshold $\varepsilon$ defined in Equation \ref{eq:fitness-penalty} as the max regret returned by RegretNet during training (the lowest one across different parameter initializations), which is typically small as demonstrated in \cite{golowich2018deep}. 

For LLMMech-e, we set the size of top individuals in the heuristic population, $b$, to 3; the number of maximum consecutive generations without improvement (for detecting stagnation), $T_p$, to 3; the size of the top individuals produced by a prompt strategy  (for measuring its fitness), $d$, to 3; and the size of the prompt population to 5.
During prompt evolution, we notice more complex prompt strategies may return highly-inefficient mechanisms.
Therefore, for quality control of the LLM-generated prompts, we only admit one to the prompt strategy population if at least one of its produced mechanisms is valid, i.e., not experiencing timeout.

All above choices were made mainly to ensure reasonable runtime (typically half an hour for LLMMech and a third quarter to an hour for LLMMech-e) without hampering the generation of nontrivial mechanisms.
We believe the presented results for LLM methods can be further improved if we increase the scale of evolution.

\subsection{Main Results}\label{subapp:res}


\paragraph{Weighted Social Cost with Arbitrary Weights.}
We report the sets of arbitrary weights used in Table \ref{tab:arbi-weights}.
They were sampled uniformly from $\{1,\ldots,5\}$.
\begin{table}[h]
    \footnotesize
    \centering
    \begin{tabular}{|c|c|c|}
        \hline
        \# & $n=5$ & $n=10$ \\\hline
        1 & 5 3 3 3 1 & 5 4 4 3 2 2 2 1 1 1\\
        2 & 4 4 4 3 3 & 5 5 4 3 2 2 2 1 1 1\\
        3 & 3 3 3 2 1 & 5 5 4 3 3 2 2 2 1 1\\
        4 & 5 2 1 1 1 & 5 5 4 4 4 3 2 2 1 1\\
        5 & 5 4 3 3 2 & 4 4 4 4 4 4 4 3 2 2\\
        6 & 4 4 4 3 1 & 5 4 4 3 3 2 1 1 1 1\\
        7 & 4 2 2 2 1 & 5 5 4 3 3 3 3 2 2 1\\
        8 & 5 4 2 2 1 & 5 4 4 3 3 3 3 2 2 1\\
        9 & 5 4 4 3 2 & 5 5 5 5 4 4 4 4 4 4\\
        10 & 5 4 4 3 3 & 5 5 5 4 4 3 2 2 2 1\\\hline
    \end{tabular}
    \caption{Set of arbitrary weights used.}
    \label{tab:arbi-weights}
\end{table}

\begin{figure*}
    \centering
    \begin{subfigure}[t]{0.99\textwidth}
        \centering
        \includegraphics[width=\textwidth]{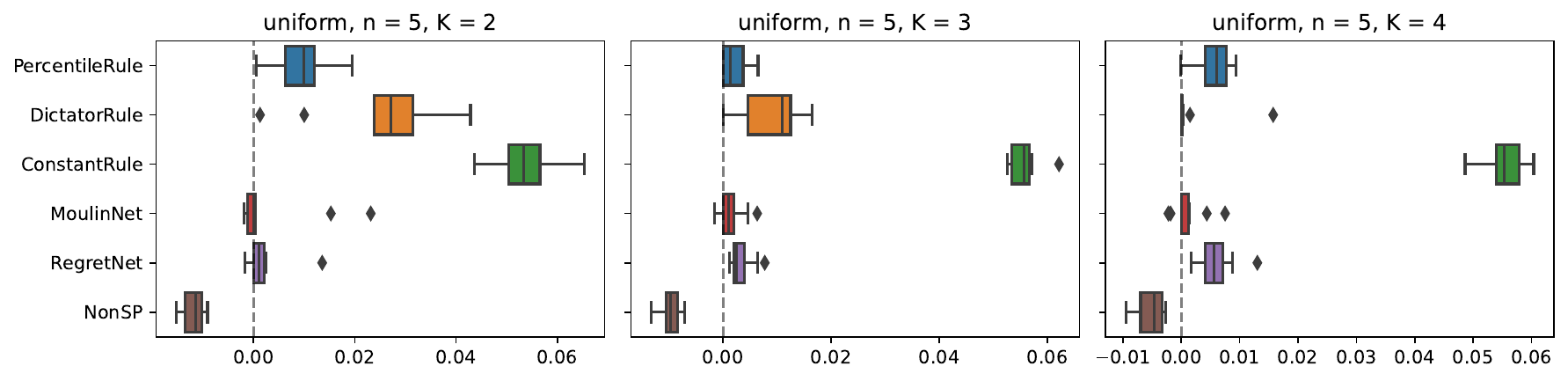}
    \end{subfigure}\\
    \begin{subfigure}[t]{0.99\textwidth}
        \centering
        \includegraphics[width=\textwidth]{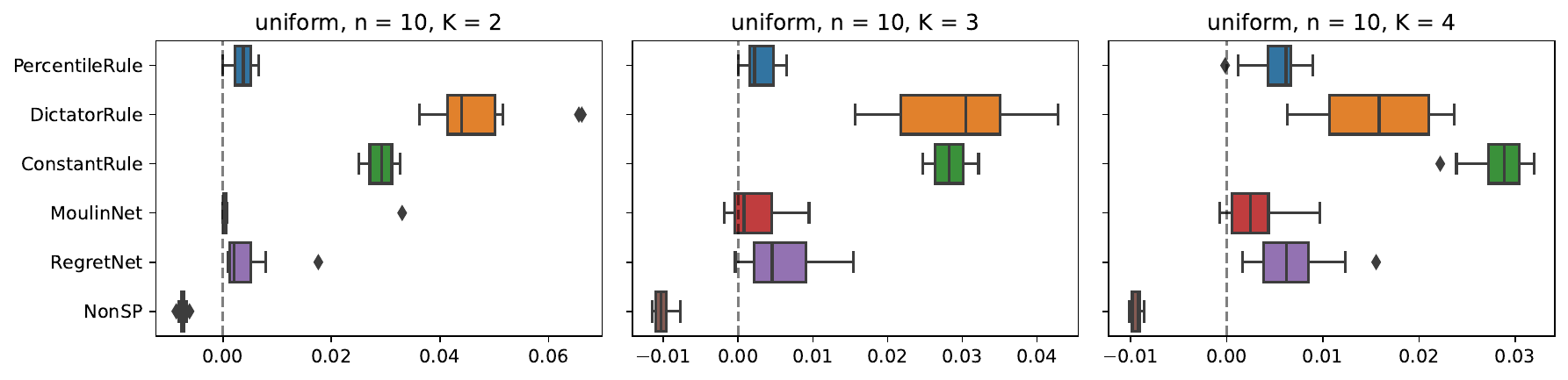}
    \end{subfigure}\\
    \begin{subfigure}[t]{0.99\textwidth}
        \centering
        \includegraphics[width=\textwidth]{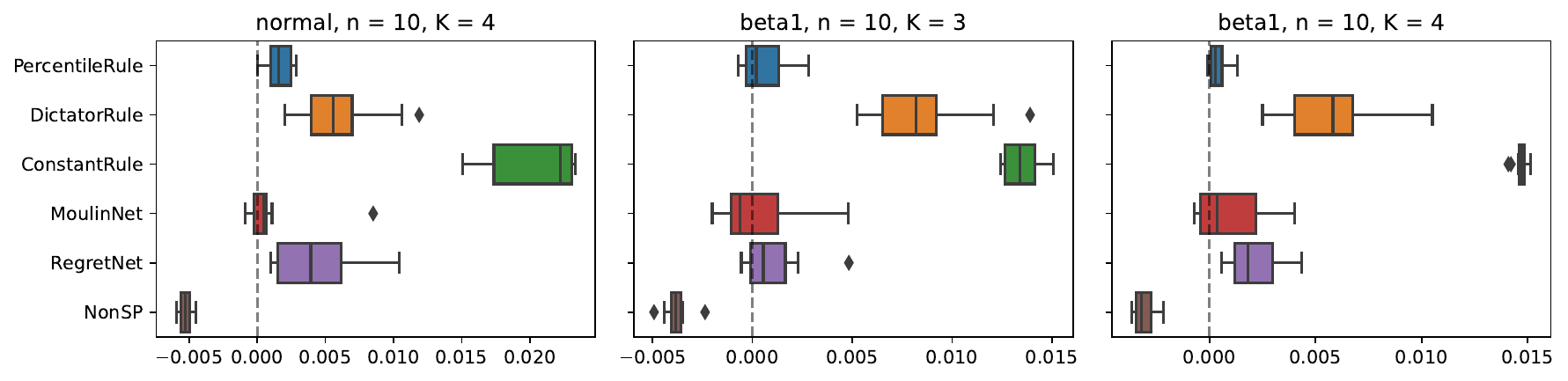}
    \end{subfigure}\\
    \begin{subfigure}[t]{0.66\textwidth}
        \centering
        \includegraphics[width=\textwidth]{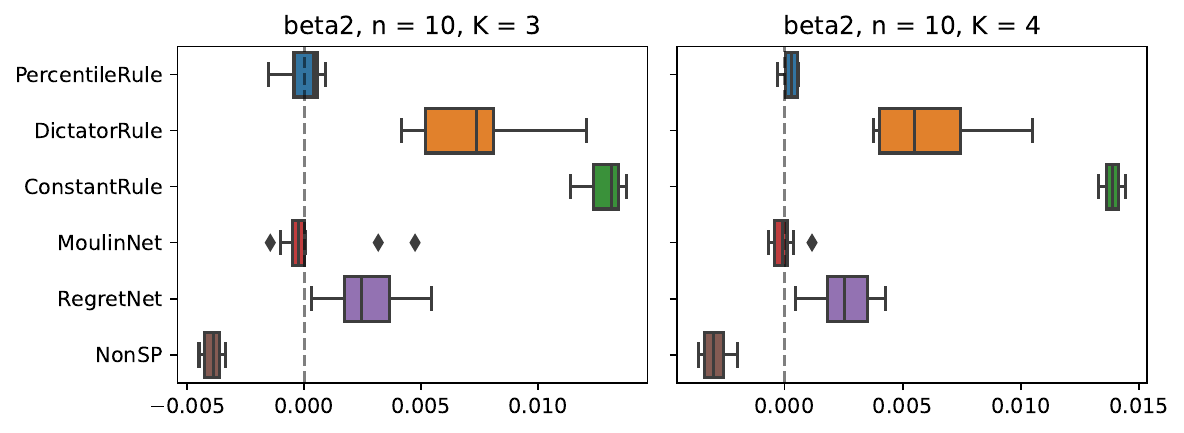}
    \end{subfigure}
    \caption{Results for net difference with respect to LLMMech-e in (arbitrarily) weighted social cost.}
    \label{fig:res-aw}
\end{figure*}

\begin{figure*}
    \centering
    \begin{subfigure}[t]{0.99\textwidth}
        \centering
        \includegraphics[width=\textwidth]{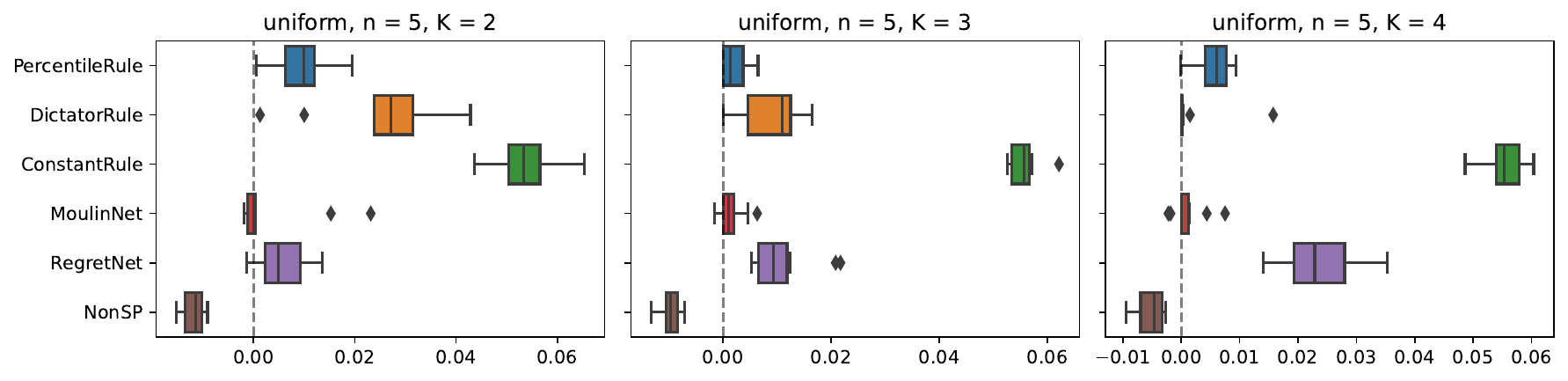}
    \end{subfigure}\\
    \begin{subfigure}[t]{0.99\textwidth}
        \centering
        \includegraphics[width=\textwidth]{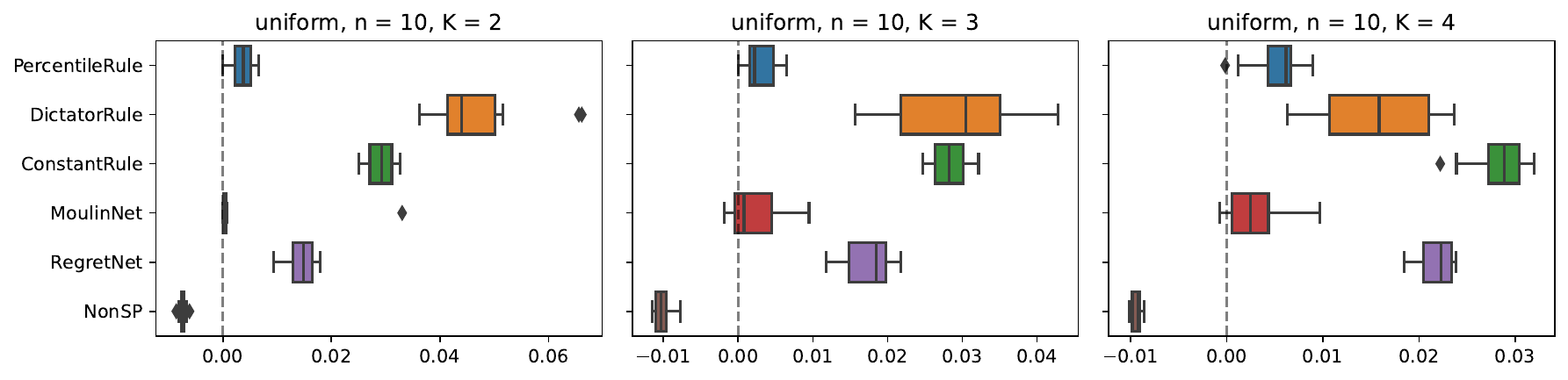}
    \end{subfigure}\\
    \begin{subfigure}[t]{0.99\textwidth}
        \centering
        \includegraphics[width=\textwidth]{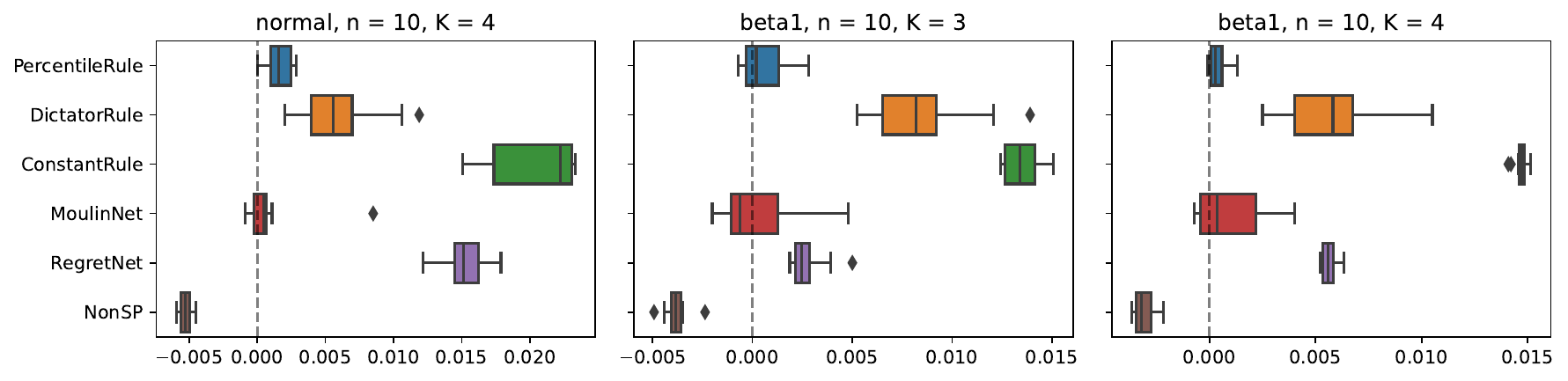}
    \end{subfigure}\\
    \begin{subfigure}[t]{0.66\textwidth}
        \centering
        \includegraphics[width=\textwidth]{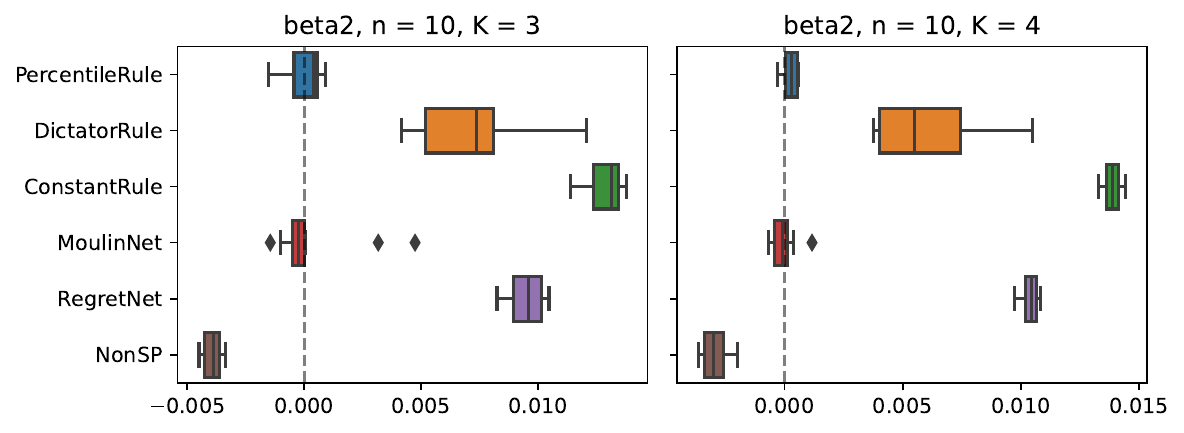}
    \end{subfigure}
    \caption{(Old) Results for net difference with respect to LLMMech-e in (arbitrarily) weighted social cost. RegretNet was trained on 25k samples without hyperparameter tuning of batch size and $\rho$.}
    \label{fig:res-aw-old}
\end{figure*}

Figure \ref{fig:res-aw} shows results for $n=5$ and 10. 
Figure \ref{fig:res-aw-old} shows similar results but with untuned RegretNet.

\clearpage
\paragraph{Unweighted Social Cost.}

Tables \ref{tab:res-uw-n5}, \ref{tab:res-uw-n10}, \ref{tab:res-uw-n25} respectively show results for $n=5,10,25$, each with $K\in\{1,2,3,4\}$ (we additionally consider $K=5$ for $n=25$).

\begin{table*}[h]
    \footnotesize
    \centering
    \begin{tabular}{|ccccc|ccllc|}
        \hline
        & $K$ & Per. & Dict. & Cons. & MoulinNet & RegretNet & \multicolumn{1}{c}{LLMMech} & \multicolumn{1}{c}{LLMMech-e} & NonSP \\\hline
        uniform & 1 & \textbf{0.20038} & 0.26673 & 0.25242 & 0.20045 & 0.21570 (0.00327) & \textbf{0.20038} & \textbf{0.20038} & 0.20038 \\
        uniform & 2 & 0.08399 & 0.12589 & 0.12708 & 0.10371 & 0.09599 (0.00292) & 0.08398 & \textbf{0.08373} & 0.07098 \\
        uniform & 3 & 0.03328 & 0.06038 & 0.09020 & 0.04027 & 0.05295 (0.00310) & 0.03328 & \textbf{0.03323} (0.00017) & 0.02798 \\
        uniform & 4 & 0.01679 & 0.02324 & 0.06552 & 0.01689 & 0.04630 (0.00283) & \textbf{0.00852} (0.00070) & \textbf{0.00852} (0.00070) & 0.00852 \\\hline
        normal & 1 & \textbf{0.09444} & 0.12805 & 0.11955 & 0.09448 & 0.10137 (0.00168) & \textbf{0.09444} & \textbf{0.09444} & 0.09444 \\
        normal & 2 & 0.04443 & 0.06138 & 0.07134 & 0.04445 & 0.05231 (0.00131) & \textbf{0.04343} & \textbf{0.04343} & 0.03729 \\
        normal & 3 & 0.01698 & 0.03072 & 0.05390 & 0.02049 & 0.02755 (0.00102) & 0.01698 & \textbf{0.01478} (0.00177) & 0.01469 \\
        normal & 4 & 0.00715 & 0.01202 & 0.03846 & 0.00835 & 0.02128 (0.00111) & \textbf{0.00444} (0.00063) & \textbf{0.00444} (0.00063) & 0.00444 \\\hline
        beta1 & 1 & \textbf{0.05431} & 0.07342 & 0.07210 & 0.05433 & 0.05538 (0.00196) & \textbf{0.05431} & \textbf{0.05431} & 0.05431 \\
        beta1 & 2 & \textbf{0.02013} & 0.03454 & 0.04765 & \textbf{0.02013} & 0.02286 (0.00135) & \textbf{0.02013} & \textbf{0.02013} & 0.01875 \\
        beta1 & 3 & 0.00912 & 0.01686 & 0.03207 & 0.01186 & 0.01167 (0.00173) & \textbf{0.00672} (0.00068) & \textbf{0.00672} (0.00068) & 0.00672 \\
        beta1 & 4 & 0.00287 & 0.00675 & 0.02812 & 0.00340 & 0.00952 (0.00146) & \textbf{0.00198} (0.00017) & \textbf{0.00198} (0.00017) & 0.00198 \\\hline
        beta2 & 1 & \textbf{0.05530} & 0.07597 & 0.07273 & 0.05533 & 0.06389 (0.00207) & \textbf{0.05530} & \textbf{0.05530} & 0.05530 \\
        beta2 & 2 & \textbf{0.02023} & 0.03425 & 0.04839 & \textbf{0.02023} & 0.03717 (0.00115) & 0.02455 & \textbf{0.02023} & 0.01877 \\
        beta2 & 3 & 0.00903 & 0.01735 & 0.03234 & 0.01215 & 0.02309 (0.00053) & \textbf{0.00667} (0.00081) & \textbf{0.00667} (0.00081) & 0.00667 \\
        beta2 & 4 & 0.00299 & 0.00708 & 0.02830 & 0.00302 & 0.01956 (0.00015) & 0.00299 & \textbf{0.00293} (0.00005) & 0.00195 \\\hline
    \end{tabular}
    \caption{Results for $n=5$, unweighted social cost. Values in parentheses (if any) represent regret (max across agents) incurred by the mechanism. The best results (beside NonSP) are bolded.}
    \label{tab:res-uw-n5}
\end{table*}

\begin{table*}[h]
    \footnotesize
    \centering
    \begin{tabular}{|ccc|cc|ll|c|}
        \hline
        & $K$ & Per. & RegretNet & RegretNet@25k & \multicolumn{1}{c}{LLMMech} & \multicolumn{1}{c|}{LLMMech-e} & NonSP \\\hline
        uniform & 1 & \textbf{0.20038} & 0.21570 (0.00327) & 0.20128 (0.00126) & \textbf{0.20038} & \textbf{0.20038} & 0.20038 \\
        uniform & 2 & 0.08399 & 0.09599 (0.00292) & 0.08501 (0.00290) & 0.08399 & \textbf{0.08373} & 0.07098 \\
        uniform & 3 & 0.03328 & 0.05295 (0.00310) & 0.03493 (0.00206) & 0.03327 & \textbf{0.03323} (0.00017) & 0.02798 \\
        uniform & 4 & 0.01679 & 0.04630 (0.00283) & 0.01999 (0.00156) & \textbf{0.00852} (0.00070) & \textbf{0.00852} (0.00070) & 0.00852 \\\hline
    \end{tabular}
    \caption{(Additional) Results for $n=5$, unweighted social cost. `RegretNet@25k' denotes RegretNet when trained on 25k samples. Values in parentheses (if any) represent regret (max across agents) incurred by the mechanism. The best results (beside NonSP) are bolded.}
    \label{tab:res-uw-n5-extra}
\end{table*}

\begin{table*}[h]
    \footnotesize
    \centering
    \begin{tabular}{|ccccc|clllc|}
        \hline
        & $K$ & Per. & Dict. & Cons. & MoulinNet & \multicolumn{1}{c}{RegretNet} & \multicolumn{1}{c}{LLMMech} & \multicolumn{1}{c}{LLMMech-e} & NonSP \\\hline
        uniform & 1 & \textbf{0.22759} & 0.29938 & 0.25120 & 0.22764 & 0.24228 (0.00264) & \textbf{0.22759} & \textbf{0.22759} & 0.22759 \\
        uniform & 2 & 0.10004 & 0.16933 & 0.12463 & 0.10009 & 0.11796 (0.00310) & 0.10004 & \textbf{0.09979} (0.00024) & 0.09458 \\
        uniform & 3 & 0.05922 & 0.10396 & 0.08410 & 0.05926 & 0.07993 (0.00201) & \textbf{0.05892} & \textbf{0.05892} & 0.05196 \\
        uniform & 4 & 0.04096 & 0.06973 & 0.06340 & 0.04899 & 0.06046 (0.00221) & \textbf{0.04075} & \textbf{0.04075} & 0.03173 \\\hline
        normal & 1 & \textbf{0.09585} & 0.13054 & 0.11977 & \textbf{0.09585} & 0.10504 (0.00102) & \textbf{0.09585} & \textbf{0.09585} & 0.09585 \\
        normal & 2 & \textbf{0.04955} & 0.07612 & 0.08200 & 0.04981 & 0.06361 (0.00236) & \textbf{0.04955} & \textbf{0.04955} & 0.04629 \\
        normal & 3 & 0.03067 & 0.04828 & 0.04949 & 0.03091 & 0.04786 (0.00187) & \textbf{0.03045} & \textbf{0.03045} & 0.02651 \\
        normal & 4 & 0.02042 & 0.03351 & 0.04092 & 0.02136 & 0.03611 (0.00119) & \textbf{0.02005} & \textbf{0.02005} & 0.01594 \\\hline
        beta1 & 1 & \textbf{0.06310} & 0.08564 & 0.07052 & 0.06311 & 0.06436 (0.00192) & \textbf{0.06310} & \textbf{0.06310} & 0.06310 \\
        beta1 & 2 & \textbf{0.03022} & 0.04722 & 0.04006 & 0.03056 & 0.03483 (0.00292) & 0.03078 & \textbf{0.03022} & 0.02794 \\
        beta1 & 3 & \textbf{0.01703} & 0.03008 & 0.03063 & 0.01726 & 0.02251 (0.00252) & 0.01892 & 0.01769 & 0.01493 \\
        beta1 & 4 & 0.01029 & 0.02196 & 0.02505 & 0.01047 & 0.01638 (0.00154) & 0.01029 & \textbf{0.00949} (0.00103) & 0.00866 \\\hline
        beta2 & 1 & \textbf{0.06092} & 0.08275 & 0.07048 & \textbf{0.06092} & 0.06411 (0.00246) & \textbf{0.06092} & \textbf{0.06092} & 0.06092 \\
        beta2 & 2 & 0.02934 & 0.04604 & 0.03903 & 0.03243 & 0.03764 (0.00140) & \textbf{0.02720} (0.00169) & 0.02989 & 0.02720 \\
        beta2 & 3 & \textbf{0.01668} & 0.03074 & 0.02976 & 0.01670 & 0.02641 (0.00042) & 0.01850 & \textbf{0.01668} & 0.01461 \\
        beta2 & 4 & \textbf{0.01031} & 0.02103 & 0.02402 & 0.01239 & 0.02070 & 0.01088 & 0.01088 & 0.00847 \\\hline
    \end{tabular}
    \caption{Results for $n=10$, unweighted social cost. Values in parentheses (if any) represent regret (max across agents) incurred by the mechanism. The best results (beside NonSP) are bolded.}
    \label{tab:res-uw-n10}
\end{table*}

\begin{table*}[h]
    \footnotesize
    \centering
    \begin{tabular}{|ccccc|ccccc|}
        \hline
        & $K$ & Per. & Dict. & Cons. & MoulinNet & RegretNet & LLMMech & LLMMech-e & NonSP \\\hline
        uniform & 1 & \textbf{0.24099} & 0.32248 & 0.25120 & 0.24101 & 0.24814 (0.00263) & \textbf{0.24099} & \textbf{0.24099} & 0.24099 \\
        uniform & 2 & \textbf{0.11636} & 0.19376 & 0.12631 & 0.11665 & 0.12503 (0.00250) & \textbf{0.11636} & \textbf{0.11636} & 0.11248 \\
        uniform & 3 & \textbf{0.07406} & 0.13425 & 0.08389 & 0.07456 & 0.08340 (0.00229) & \textbf{0.07406} & \textbf{0.07406} & 0.06857 \\
        uniform & 4 & \textbf{0.05294} & 0.09788 & 0.06301 & 0.05352 & 0.06262 (0.00228) & 0.05296 & 0.05296 & 0.04736 \\
        uniform & 5 & 0.04070 & 0.07486 & 0.05075 & 0.04041 & 0.04994 (0.00239) & \textbf{0.03990} & \textbf{0.03990} & 0.03502 \\\hline
        normal & 1 & \textbf{0.08969} & 0.12597 & 0.09423 & 0.08977 & 0.09147 (0.00265) & \textbf{0.08969} & \textbf{0.08969} & 0.08969 \\
        normal & 2 & 0.05094 & 0.07519 & 0.05766 & 0.05112 & 0.05520 (0.00218) & \textbf{0.05089} & 0.05094 & 0.04904 \\
        normal & 3 & 0.03482 & 0.05456 & 0.04016 & 0.03473 & 0.03988 (0.00184) & 0.03525 & \textbf{0.03453} & 0.03213 \\
        normal & 4 & \textbf{0.02575} & 0.04212 & 0.03218 & 0.02606 & 0.03174 (0.00111) & 0.02631 & \textbf{0.02575} & 0.02291 \\
        normal & 5 & \textbf{0.02030} & 0.03460 & 0.02810 & 0.02112 & 0.02681 (0.00105) & 0.02087 & 0.02032 & 0.01706 \\\hline
        beta1 & 1 & \textbf{0.06439} & 0.08979 & 0.06687 & 0.06440 & 0.06517 (0.00199) & \textbf{0.06439} & \textbf{0.06439} & 0.06439 \\
        beta1 & 2 & \textbf{0.03440} & 0.05356 & 0.03771 & 0.03452 & 0.03696 (0.00296) & 0.03496 & \textbf{0.03440} & 0.03309 \\
        beta1 & 3 & \textbf{0.02279} & 0.03830 & 0.02701 & 0.02293 & 0.02598 (0.00277) & 0.02379 & 0.02340 & 0.02073 \\
        beta1 & 4 & \textbf{0.01622} & 0.02971 & 0.02097 & 0.01795 & 0.02002 (0.00296) & 0.01707 & 0.01689 & 0.01430 \\
        beta1 & 5 & 0.01235 & 0.02421 & 0.01832 & \textbf{0.01231} & 0.01639 (0.00281) & 0.01327 & 0.01327 & 0.01036 \\\hline
        beta2 & 1 & \textbf{0.06434} & 0.09241 & 0.06675 & 0.06435 & 0.06555 (0.00145) & \textbf{0.06434} & \textbf{0.06434} & 0.06434 \\
        beta2 & 2 & \textbf{0.03466} & 0.05319 & 0.03812 & 0.03485 & 0.03782 (0.00056) & 0.03524 & 0.03512 & 0.03327 \\
        beta2 & 3 & \textbf{0.02305} & 0.03816 & 0.02712 & 0.02467 & 0.02671 (0.00000) & 0.02410 & 0.02341 & 0.02083 \\
        beta2 & 4 & \textbf{0.01658} & 0.03053 & 0.02234 & 0.01667 & 0.02056 (0.00000) & 0.01744 & 0.01670 & 0.01432 \\
        beta2 & 5 & \textbf{0.01230} & 0.02449 & 0.01775 & 0.01409 & 0.01670 (0.00000) & 0.01371 & 0.01322 & 0.01041 \\\hline
    \end{tabular}
    \caption{Results for $n=25$, unweighted social cost. Values in parentheses (if any) represent regret (max across agents) incurred by the mechanism. The best results (beside NonSP) are bolded.}
    \label{tab:res-uw-n25}
\end{table*}

\clearpage
\subsection{Another Case Study on Learned Mechanisms.}\label{subapp:case-study}
Figure \ref{fig:learned-mech2} provides another example of a learned mechanism from LLMMech-e. 
It yields second lowest cost (behind vanilla LLMMech, and with zero empirical regret) compared to other baselines for the setting of $n=10,K=4$ with normal peak distribution, as reported in Table \ref{tab:res-w-n5+10}.
Essentially, the mechanism first transforms the larger agent weights into repeated occurrences of agents (i.e., the agent with weight 5 will be transformed to five agents with weight 1 at the same location), then applies the percentile rule as in the unweighted case.
Because the agents cannot misreport their weights, the mechanism is strategyproof even for the general case. 
Here, compared to the na\"{i}ve generalization of the percentile mechanism to weighted settings\footnote{In \cite{golowich2018deep}'s and our generalization, the agent weights are incorporated \emph{post hoc} during the computation of social cost for each combination of percentiles.}, the problem transformation step is novel.

\begin{figure}[h]
    \footnotesize
    \centering
    \begin{tcolorbox}[width=0.65\textwidth, colback=blue!5]
        \textbf{Mechanism Description}

        \hphantom\\

        The new strategy involves using the K-means clustering algorithm to form clusters from the location samples and selecting the centroid of each cluster as the facility locations to balance the distribution of utility and minimize the total cost.
    \end{tcolorbox}
    \begin{lstlisting}[captionpos=b]
def get_locations(samples):
    '''
    Determines the optimal locations from a given list of location samples.

    Args:
    samples (list): A one-dimensional list containing the location samples.
    weights (list): A list of fixed weights assigned to the samples: [5,5,1,1,1,1,1,1,1,1]

    Returns:
    list: A one-dimensional list of the optimal locations, containing n_locations elements in [0,1].
    '''
    n_locations = 4
    weights = [5, 5, 1, 1, 1, 1, 1, 1, 1, 1]

    # Combine samples and weights into a single list based on weights
    weighted_samples = sorted(zip(samples, weights))
    sample_weighted = [sample for sample, weight in weighted_samples for _ in range(weight)]

    # Perform K-means clustering by selecting centroids of grouped samples
    step = len(sample_weighted) // n_locations + (len(sample_weighted) % n_locations > 0)
    locations = [sample_weighted[i * step + step // 2] for i in range(n_locations)]

    return locations
    \end{lstlisting}
    \caption{(Top) High-level description of a learned mechanism and (bottom) its associated code.}
    \label{fig:learned-mech2}
\end{figure}

\clearpage
\subsection{Ablation Study}\label{subapp:ablation}

\paragraph{Out-of-Distribution Performance (Weighted).}

Tables \ref{tab:res-w-n5-ood} and \ref{tab:res-w-n10-ood} respectively show results for
weighted $n=5$ and 10, each with $K\in\{1,2,3,4\}$.

\begin{table*}[h]
    \footnotesize
    \centering
    \begin{tabular}{|cccccc|cl|c|}
        \hline
        & $K$ & Per. & Dict. & MoulinNet & RegretNet@25k & LLMMech (OOD) & \multicolumn{1}{c|}{LLMMech-e (OOD)} & NonSP \\\hline
        normal & 1 & 0.09350 & \textbf{0.07148} & 0.07150 & 0.07178 (0.00021) & \textbf{0.07148} & \textbf{0.07148} & 0.07148 \\
        normal & 2 & 0.04302 & 0.03360 & 0.03402 & 0.03503 (0.00038) & 0.03097 & \textbf{0.02760} (0.00027) & 0.02303 \\
        normal & 3 & 0.01648 & 0.01688 & 0.01624 & 0.02348 (0.00030) & 0.01628 & \textbf{0.01040} (0.00235) & 0.00865 \\
        normal & 4 & 0.00823 & 0.00694 & 0.00569 & 0.01902 (0.00015) & 0.00699 & \textbf{0.00448} (0.00015) & 0.00247 \\\hline
        beta1 & 1 & 0.05422 & \textbf{0.04079} & \textbf{0.04079} & 0.04094 (0.00009) & \textbf{0.04079} & \textbf{0.04079} & 0.04079 \\
        beta1 & 2 & 0.02396 & 0.01904 & 0.01575 & 0.01838 (0.00033) & 0.01861 & \textbf{0.01508} (0.00026) & 0.01147 \\
        beta1 & 3 & 0.00866 & 0.00978 & 0.00655 & 0.00928 (0.00044) & 0.00850 & \textbf{0.00604} (0.00151) & 0.00400 \\
        beta1 & 4 & 0.00305 & 0.00378 & \textbf{0.00188} & 0.00617 (0.00048) & 0.00390 & 0.00194 (0.00007) & 0.00110 \\\hline
        beta2 & 1 & 0.06057 & \textbf{0.04128} & \textbf{0.04128} & 0.04166 (0.00007) & \textbf{0.04128} & \textbf{0.04128} & 0.04128 \\
        beta2 & 2 & 0.01978 & 0.01918 & 0.01571 & 0.01957 (0.00012) & 0.01683 & \textbf{0.01324} & 0.01164 \\
        beta2 & 3 & 0.00856 & 0.00980 & 0.00666 & 0.01449 (0.00010) & 0.00792 & \textbf{0.00555} & 0.00393 \\
        beta2 & 4 & 0.00305 & 0.00407 & \textbf{0.00189} & 0.01671 (0.00013) & 0.00392 & 0.00203 (0.00007) & 0.00108 \\\hline
    \end{tabular}
    \begin{tabular}{|cccccc|cl|c|}
        \hline
        & $K$ & Per. (OOD) & Dict. (OOD) & MoulinNet (OOD) & RegretNet@25k (OOD) & LLMMech (OOD) & \multicolumn{1}{c|}{LLMMech-e (OOD)} & NonSP \\\hline
        normal & 1 & \textbf{0.07148} & \textbf{0.07148} & \textbf{0.07148} & 0.07164 (0.00019) & \textbf{0.07148} & \textbf{0.07148} & 0.07148 \\
        normal & 2 & 0.03410 & 0.03410 & 0.02821 & 0.04637 (0.00057) & 0.03097 & \textbf{0.02760} (0.00027) & 0.02303 \\
        normal & 3 & 0.01688 & 0.01688 & 0.01646 & 0.03901 (0.00053) & 0.01628 & \textbf{0.01040} (0.00235) & 0.00865 \\
        normal & 4 & 0.00694 & 0.00694 & 0.00451 & 0.02619 (0.00033) & 0.00699 & \textbf{0.00448} (0.00015) & 0.00247 \\\hline
        beta1 & 1 & \textbf{0.04079} & \textbf{0.04079} & \textbf{0.04079} & 0.04150 (0.00031) & \textbf{0.04079} & \textbf{0.04079} & 0.04079 \\
        beta1 & 2 & 0.01932 & 0.01932 & 0.01675 & 0.04012 (0.00024) & 0.01861 & \textbf{0.01508} (0.00026) & 0.01147 \\
        beta1 & 3 & 0.00939 & 0.00939 & 0.01036 & 0.03473 (0.00037) & 0.00850 & \textbf{0.00604} (0.00151) & 0.00400 \\
        beta1 & 4 & 0.00394 & 0.00394 & 0.00494 & 0.03227 (0.00030) & 0.00390 & \textbf{0.00194} (0.00007) & 0.00110 \\\hline
        beta2 & 1 & \textbf{0.04128} & \textbf{0.04128} & \textbf{0.04128} & 0.04183 (0.00026) & \textbf{0.04128} & \textbf{0.04128} & 0.04128 \\
        beta2 & 2 & 0.01925 & 0.01925 & 0.01825 & 0.04037 (0.00036) & 0.01683 & \textbf{0.01324} & 0.01164 \\
        beta2 & 3 & 0.00964 & 0.00964 & 0.00870 & 0.03176 (0.00035) & 0.00792 & \textbf{0.00555} & 0.00393 \\
        beta2 & 4 & 0.00381 & 0.00381 & 0.00303 & 0.02388 (0.00028) & 0.00392 & \textbf{0.00203} (0.00007) & 0.00108 \\\hline
    \end{tabular}
    \caption{Out-of-distribution (OOD) results for $n=5$, weighted social cost. Baselines without `OOD' were trained with in-distribution (uniformly distributed) samples. Values in parentheses (if any) represent regret (max across agents) incurred by the mechanism. The best results (beside NonSP) are bolded.}
    \label{tab:res-w-n5-ood}
\end{table*}

\begin{table*}[h]
    \footnotesize
    \centering
    \begin{tabular}{|cccccl|cl|c|}
        \hline
        & $K$ & Per. & Dict. & MoulinNet & \multicolumn{1}{c|}{RegretNet@25k} & LLMMech (OOD) & \multicolumn{1}{c|}{LLMMech-e (OOD)} & NonSP \\\hline
        normal & 1 & 0.09537 & 0.10546 & 0.08795 & 0.09244 (0.00029) & \textbf{0.08791} & \textbf{0.08791} & 0.08791 \\
        normal & 2 & 0.04909 & 0.04124 & 0.03847 & 0.05340 (0.00060) & \textbf{0.03802} & \textbf{0.03802} & 0.03489 \\
        normal & 3 & 0.03041 & 0.02729 & 0.02485 & 0.02800 (0.00061) & 0.02310 & \textbf{0.02228} & 0.01779 \\
        normal & 4 & 0.02051 & 0.01927 & 0.01674 & 0.02252 (0.00044) & 0.01822 & \textbf{0.01471} (0.00053) & 0.01012 \\\hline
        beta1 & 1 & 0.06260 & 0.06851 & 0.05781 & 0.05848 (0.00069) & \textbf{0.05780} & \textbf{0.05780} & 0.05780 \\
        beta1 & 2 & 0.03044 & 0.02677 & 0.02380 & 0.02378 (0.00085) & 0.02405 & \textbf{0.02361} (0.00070) & 0.02107 \\
        beta1 & 3 & 0.01720 & 0.01762 & \textbf{0.01474} & 0.01571 (0.00047) & 0.01485 & 0.01475 & 0.01025 \\
        beta1 & 4 & 0.01035 & 0.01259 & \textbf{0.00799} & 0.01190 (0.00077) & 0.01227 & 0.00949 & 0.00562 \\\hline
        beta2 & 1 & 0.06073 & 0.06644 & \textbf{0.05617} & 0.06502 (0.00122) & \textbf{0.05617} & \textbf{0.05617} & 0.05617 \\
        beta2 & 2 & 0.03120 & 0.02538 & 0.02319 & 0.03777 (0.00070) & 0.02342 & \textbf{0.02289} (0.00041) & 0.02060 \\
        beta2 & 3 & 0.01663 & 0.01683 & 0.01381 & 0.02642 (0.00000) & \textbf{0.01294} & 0.01438 & 0.01000 \\
        beta2 & 4 & 0.01016 & 0.01177 & 0.00883 & 0.02055 & 0.01179 & \textbf{0.00799} (0.00026) & 0.00549 \\\hline
    \end{tabular}
    \begin{tabular}{|cccccc|cl|c|}
        \hline
        & $K$ & Per. (OOD) & Dict. (OOD) & MoulinNet (OOD) & RegretNet@25k (OOD) & LLMMech (OOD) & \multicolumn{1}{c|}{LLMMech-e (OOD)} & NonSP \\\hline
        normal & 1 & 0.10546 & 0.10546 & \textbf{0.08791} & 0.09213 (0.00201) & \textbf{0.08791} & \textbf{0.08791} & 0.08791 \\
        normal & 2 & 0.04124 & 0.04124 & 0.03827 & 0.04568 (0.00174) & \textbf{0.03802} & \textbf{0.03802} & 0.03489 \\
        normal & 3 & 0.02716 & 0.02716 & 0.02343 & 0.04154 (0.00174) & 0.02310 & \textbf{0.02228} & 0.01779 \\
        normal & 4 & 0.01881 & 0.01881 & 0.01494 & 0.03180 (0.00047) & 0.01822 & \textbf{0.01471} (0.00053) & 0.01012 \\\hline
        beta1 & 1 & 0.06868 & 0.06868 & 0.05939 & 0.07546 (0.00335) & \textbf{0.05780} & \textbf{0.05780} & 0.05780 \\
        beta1 & 2 & 0.02677 & 0.02677 & 0.02393 & 0.05450 (0.00537) & 0.02405 & \textbf{0.02361} (0.00070) & 0.02107 \\
        beta1 & 3 & 0.01755 & 0.01755 & 0.01542 & 0.02808 (0.00091) & 0.01485 & \textbf{0.01475} & 0.01025 \\
        beta1 & 4 & 0.01259 & 0.01259 & 0.01003 & 0.02665 (0.00221) & 0.01227 & \textbf{0.00949} & 0.00562 \\\hline
        beta2 & 1 & 0.06802 & 0.06802 & 0.05692 & 0.07376 (0.00238) & \textbf{0.05617} & \textbf{0.05617} & 0.05617 \\
        beta2 & 2 & 0.02538 & 0.02538 & 0.02538 & 0.05442 (0.00452) & 0.02342 & \textbf{0.02289} (0.00041) & 0.02060 \\
        beta2 & 3 & 0.01713 & 0.01713 & 0.01628 & 0.02675 (0.00097) & \textbf{0.01294} & 0.01438 & 0.01000 \\
        beta2 & 4 & 0.01188 & 0.01188 & 0.00932 & 0.02641 (0.00125) & 0.01179 & \textbf{0.00799} (0.00026) & 0.00549 \\\hline
    \end{tabular}
    \caption{Out-of-distribution (OOD) results for $n=10$, weighted social cost. Baselines without `OOD' were trained with in-distribution (uniformly distributed) samples. Values in parentheses (if any) represent regret (max across agents) incurred by the mechanism. The best results (beside NonSP) are bolded.}
    \label{tab:res-w-n10-ood}
\end{table*}

\clearpage
\paragraph{Out-of-Distribution Performance (Unweighted).}
Tables \ref{tab:res-uw-n5-ood}, \ref{tab:res-uw-n10-ood}, \ref{tab:res-uw-n25-ood} respectively show results for
unweighted $n=5,10,25$, each with $K\in\{1,2,3,4\}$ (we additionally consider $K=5$ for $n=25$).

\begin{table*}[h]
    \footnotesize
    \centering
    \begin{tabular}{|ccccc|ll|c|}
        \hline
        & $K$ & Per. & MoulinNet & RegretNet@25k & \multicolumn{1}{c}{LLMMech (OOD)} & \multicolumn{1}{c|}{LLMMech-e (OOD)} & NonSP \\\hline
        normal & 1 & \textbf{0.09444} & 0.09448 & 0.10137 (0.00168) & \textbf{0.09444} & \textbf{0.09444} & 0.09444 \\
        normal & 2 & 0.04443 & 0.04445 & 0.05231 (0.00131) & \textbf{0.04343} & \textbf{0.04343} & 0.03729 \\
        normal & 3 & 0.01698 & 0.02049 & 0.02755 (0.00102) & 0.01698 & \textbf{0.01697} (0.00009) & 0.01469 \\
        normal & 4 & 0.00715 & 0.00835 & 0.02128 (0.00111) & \textbf{0.00444} (0.00046) & \textbf{0.00444} (0.00046) & 0.00444 \\\hline
        beta1 & 1 & \textbf{0.05431} & 0.05433 & 0.05538 (0.00196) & \textbf{0.05431} & \textbf{0.05431} & 0.05431 \\
        beta1 & 2 & \textbf{0.02013} & \textbf{0.02013} & 0.02286 (0.00135) & \textbf{0.02013} & \textbf{0.02013} & 0.01875 \\
        beta1 & 3 & 0.00912 & 0.01186 & 0.01167 (0.00173) & 0.00912 & \textbf{0.00908} (0.00037) & 0.00672 \\
        beta1 & 4 & 0.00287 & 0.00340 & 0.00952 (0.00146) & \textbf{0.00198} (0.00017) & \textbf{0.00198} (0.00017) & 0.00198 \\\hline
        beta2 & 1 & \textbf{0.05530} & 0.05533 & 0.06389 (0.00207) & \textbf{0.05530} & \textbf{0.05530} & 0.05530 \\
        beta2 & 2 & \textbf{0.02023} & \textbf{0.02023} & 0.03717 (0.00115) & 0.02770 & \textbf{0.02023} & 0.01877 \\
        beta2 & 3 & 0.00903 & 0.01215 & 0.02309 (0.00053) & 0.00879 & \textbf{0.00874} (0.00025) & 0.00667 \\
        beta2 & 4 & 0.00299 & 0.00302 & 0.01956 (0.00015) & \textbf{0.00195} (0.00020) & \textbf{0.00195} (0.00020) & 0.00195 \\\hline
    \end{tabular}
    \begin{tabular}{|ccccl|ll|c|}
        \hline
        & $K$ & Per. (OOD) & MoulinNet (OOD) & \multicolumn{1}{c|}{RegretNet@25k (OOD)} & \multicolumn{1}{c}{LLMMech (OOD)} & \multicolumn{1}{c|}{LLMMech-e (OOD)} & NonSP \\\hline
        normal & 1 & \textbf{0.09444} & \textbf{0.09444} & 0.10513 (0.00504) & \textbf{0.09444} & \textbf{0.09444} & 0.09444 \\
        normal & 2 & \textbf{0.04343} & 0.04479 & 0.05306 (0.00527) & \textbf{0.04343} & \textbf{0.04343} & 0.03729 \\
        normal & 3 & 0.01698 & 0.02178 & 0.05278 (0.00587) & 0.01698 & \textbf{0.01697} (0.00009) & 0.01469 \\
        normal & 4 & 0.00851 & 0.00819 & 0.03734 (0.00426) & 0.00444 (0.00046) & 0.00444 (0.00046) & 0.00444 \\\hline
        beta1 & 1 & \textbf{0.05431} & 0.05601 & 0.27392 & \textbf{0.05431} & \textbf{0.05431} & 0.05431 \\
        beta1 & 2 & \textbf{0.02013} & 0.02628 & 0.07794 (0.00018) & \textbf{0.02013} & \textbf{0.02013} & 0.01875 \\
        beta1 & 3 & 0.00912 & 0.01395 & 0.05936 (0.00038) & 0.00912 & \textbf{0.00908} (0.00037) & 0.00672 \\
        beta1 & 4 & 0.00287 & 0.00847 & 0.04622 (0.00462) & \textbf{0.00198} (0.00017) & \textbf{0.00198} (0.00017) & 0.00198 \\\hline
        beta2 & 1 & \textbf{0.05530} & 0.05777 & 0.16980 (0.00347) & \textbf{0.05530} & \textbf{0.05530} & 0.05530 \\
        beta2 & 2 & 0.02771 & 0.02070 & 0.08676 (0.00087) & 0.02770 & \textbf{0.02023} & 0.01877 \\
        beta2 & 3 & 0.00879 & 0.00882 & 0.05361 (0.00451) & 0.00879 & \textbf{0.00874} (0.00025) & 0.00667 \\
        beta2 & 4 & 0.00299 & 0.00587 & 0.05694 (0.00099) & \textbf{0.00195} (0.00020) & \textbf{0.00195} (0.00020) & 0.00195 \\\hline
    \end{tabular}
    \caption{Out-of-distribution (OOD) results for $n=5$, unweighted social cost. Baselines without `OOD' were trained with in-distribution (uniformly distributed) samples. Values in parentheses (if any) represent regret (max across agents) incurred by the mechanism. The best results (beside NonSP) are bolded.}
    \label{tab:res-uw-n5-ood}
\end{table*}

\begin{table*}[h]
    \footnotesize
    \centering
    \begin{tabular}{|ccccl|cl|c|}
        \hline
        & $K$ & Per. & MoulinNet & \multicolumn{1}{c|}{RegretNet@25k} & LLMMech (OOD) & \multicolumn{1}{c|}{LLMMech-e (OOD)} & NonSP \\\hline
        normal & 1 & \textbf{0.09585} & \textbf{0.09585} & 0.10504 (0.00102) & \textbf{0.09585} & \textbf{0.09585} & 0.09585 \\
        normal & 2 & 0.04955 & 0.04981 & 0.06361 (0.00236) & 0.04955 & \textbf{0.04930} (0.00030) & 0.04629 \\
        normal & 3 & 0.03067 & 0.03091 & 0.04786 (0.00187) & \textbf{0.03045} & \textbf{0.03045} & 0.02651 \\
        normal & 4 & 0.02042 & 0.02136 & 0.03611 (0.00119) & \textbf{0.02005} & \textbf{0.02005} & 0.01594 \\\hline
        beta1 & 1 & \textbf{0.06310} & 0.06311 & 0.06436 (0.00192) & \textbf{0.06310} & \textbf{0.06310} & 0.06310 \\
        beta1 & 2 & \textbf{0.03022} & 0.03056 & 0.03483 (0.00292) & 0.03247 & 0.03219 (0.00043) & 0.02794 \\
        beta1 & 3 & \textbf{0.01703} & 0.01726 & 0.02251 (0.00252) & 0.01960 & 0.01960 & 0.01493 \\
        beta1 & 4 & \textbf{0.01029} & 0.01047 & 0.01638 (0.00154) & 0.01249 & 0.01249 & 0.00866 \\\hline
        beta2 & 1 & \textbf{0.06092} & \textbf{0.06092} & 0.06411 (0.00246) & \textbf{0.06092} & \textbf{0.06092} & 0.06092 \\
        beta2 & 2 & \textbf{0.02934} & 0.03243 & 0.03764 (0.00140) & 0.03149 & 0.03115 (0.00044) & 0.02720 \\
        beta2 & 3 & \textbf{0.01668} & 0.01670 & 0.02641 (0.00042) & 0.01889 & 0.01889 & 0.01461 \\
        beta2 & 4 & \textbf{0.01031} & 0.01239 & 0.02070 & 0.01223 & 0.01223 & 0.00847 \\\hline
    \end{tabular}
    \begin{tabular}{|ccccc|cl|c|}
        \hline
        & $K$ & Per. (OOD) & MoulinNet (OOD) & RegretNet@25k (OOD) & LLMMech (OOD) & \multicolumn{1}{c|}{LLMMech-e (OOD)} & NonSP \\\hline
        normal & 1 & \textbf{0.09585} & \textbf{0.09585} & 0.10278 (0.00393) & \textbf{0.09585} & \textbf{0.09585} & 0.09585 \\
        normal & 2 & 0.04955 & 0.05847 & 0.11140 (0.00187) & 0.04955 & \textbf{0.04930} (0.00030) & 0.04629 \\
        normal & 3 & 0.03067 & 0.03176 & 0.06171 (0.00173) & \textbf{0.03045} & \textbf{0.03045} & 0.02651 \\
        normal & 4 & 0.02163 & 0.02167 & 0.06457 (0.00005) & \textbf{0.02005} & \textbf{0.02005} & 0.01594 \\\hline
        beta1 & 1 & \textbf{0.06310} & 0.06638 & 0.17242 (0.00395) & \textbf{0.06310} & \textbf{0.06310} & 0.06310 \\
        beta1 & 2 & 0.03247 & 0.03245 & 0.07132 (0.00470) & 0.03247 & \textbf{0.03219} (0.00043) & 0.02794 \\
        beta1 & 3 & 0.02073 & 0.01984 & 0.06198 (0.00380) & \textbf{0.01960} & \textbf{0.01960} & 0.01493 \\
        beta1 & 4 & 0.01666 & 0.01509 & 0.04125 (0.00590) & \textbf{0.01249} & \textbf{0.01249} & 0.00866 \\\hline
        beta2 & 1 & \textbf{0.06092} & 0.06446 & 0.20887 (0.00212) & \textbf{0.06092} & \textbf{0.06092} & 0.06092 \\
        beta2 & 2 & 0.03149 & \textbf{0.03092} & 0.12288 (0.00093) & 0.03149 & 0.03115 (0.00044) & 0.02720 \\
        beta2 & 3 & \textbf{0.01889} & 0.02027 & 0.08323 (0.00171) & \textbf{0.01889} & \textbf{0.01889} & 0.01461 \\
        beta2 & 4 & \textbf{0.01031} & 0.01572 & 0.06624 (0.00001) & 0.01223 & 0.01223 & 0.00847 \\\hline
    \end{tabular}
    \caption{Out-of-distribution (OOD) results for $n=10$, unweighted social cost. Baselines without `OOD' were trained with in-distribution (uniformly distributed) samples. Values in parentheses (if any) represent regret (max across agents) incurred by the mechanism. The best results (beside NonSP) are bolded.}
    \label{tab:res-uw-n10-ood}
\end{table*}

\begin{table*}[h]
    \footnotesize
    \centering
    \begin{tabular}{|ccccc|c|c|c|}
        \hline
        & $K$ & Per. & MoulinNet & RegretNet@25k & LLMMech (OOD) & LLMMech-e (OOD) & NonSP \\\hline
        normal & 1 & \textbf{0.08969} & 0.08977 & 0.09147 (0.00265) & \textbf{0.08969} & \textbf{0.08969} & 0.08969 \\
        normal & 2 & \textbf{0.05094} & 0.05112 & 0.05520 (0.00218) & \textbf{0.05094} & \textbf{0.05094} & 0.04904 \\
        normal & 3 & 0.03482 & \textbf{0.03473} & 0.03988 (0.00184) & 0.03488 & 0.03488 & 0.03213 \\
        normal & 4 & \textbf{0.02575} & 0.02606 & 0.03174 (0.00111) & 0.02691 & 0.02622 & 0.02291 \\
        normal & 5 & \textbf{0.02030} & 0.02112 & 0.02681 (0.00105) & 0.02087 & 0.02087 & 0.01706 \\\hline
        beta1 & 1 & \textbf{0.06439} & 0.06440 & 0.06517 (0.00199) & \textbf{0.06439} & \textbf{0.06439} & 0.06439 \\
        beta1 & 2 & \textbf{0.03440} & 0.03452 & 0.03696 (0.00296) & 0.03512 & 0.03512 & 0.03309 \\
        beta1 & 3 & \textbf{0.02279} & 0.02293 & 0.02598 (0.00277) & 0.02533 & 0.02533 & 0.02073 \\
        beta1 & 4 & \textbf{0.01622} & 0.01795 & 0.02002 (0.00296) & 0.01948 & 0.01771 & 0.01430 \\
        beta1 & 5 & 0.01235 & \textbf{0.01231} & 0.01639 (0.00281) & 0.01487 & 0.01487 & 0.01036 \\\hline
        beta2 & 1 & \textbf{0.06434} & 0.06435 & 0.06555 (0.00145) & \textbf{0.06434} & \textbf{0.06434} & 0.06434 \\
        beta2 & 2 & \textbf{0.03466} & 0.03485 & 0.03782 (0.00056) & 0.03648 & 0.03617 & 0.03327 \\
        beta2 & 3 & \textbf{0.02305} & 0.02467 & 0.02671 (0.00000) & 0.02398 & 0.02398 & 0.02083 \\
        beta2 & 4 & \textbf{0.01658} & 0.01667 & 0.02056 (0.00000) & 0.01952 & 0.01952 & 0.01432 \\
        beta2 & 5 & \textbf{0.01230} & 0.01409 & 0.01670 (0.00000) & 0.01513 & 0.01513 & 0.01041 \\\hline
    \end{tabular}
    \begin{tabular}{|ccccc|c|c|c|}
        \hline
        & $K$ & Per. (OOD) & MoulinNet (OOD) & RegretNet@25k (OOD) & LLMMech (OOD) & LLMMech-e (OOD) & NonSP \\\hline
        normal & 1 & \textbf{0.08969} & \textbf{0.08969} & 0.09373 (0.00123) & \textbf{0.08969} & \textbf{0.08969} & 0.08969 \\
        normal & 2 & \textbf{0.05090} & 0.05101 & 0.13764 (0.00132) & 0.05094 & 0.05094 & 0.04904 \\
        normal & 3 & \textbf{0.03488} & 0.03506 & 0.07419 (0.00080) & \textbf{0.03488} & \textbf{0.03488} & 0.03213 \\
        normal & 4 & \textbf{0.02622} & 0.02640 & 0.06329 (0.00063) & 0.02691 & \textbf{0.02622} & 0.02291 \\
        normal & 5 & \textbf{0.02039} & 0.02452 & 0.05026 (0.00001) & 0.02087 & 0.02087 & 0.01706 \\\hline
        beta1 & 1 & \textbf{0.06439} & 0.06472 & 0.27371 (0.00054) & \textbf{0.06439} & \textbf{0.06439} & 0.06439 \\
        beta1 & 2 & \textbf{0.03512} & 0.03534 & 0.08458 (0.00354) & \textbf{0.03512} & \textbf{0.03512} & 0.03309 \\
        beta1 & 3 & \textbf{0.02533} & 0.02459 & 0.07997 (0.00382) & \textbf{0.02533} & \textbf{0.02533} & 0.02073 \\
        beta1 & 4 & \textbf{0.01771} & 0.01833 & 0.05630 (0.00417) & 0.01948 & \textbf{0.01771} & 0.01430 \\
        beta1 & 5 & 0.01520 & \textbf{0.01438} & 0.04772 (0.00270) & 0.01487 & 0.01487 & 0.01036 \\\hline
        beta2 & 1 & \textbf{0.06434} & 0.06476 & 0.30194 (0.00117) & \textbf{0.06434} & \textbf{0.06434} & 0.06434 \\
        beta2 & 2 & 0.03648 & \textbf{0.03566} & 0.10488 (0.00095) & 0.03648 & 0.03617 & 0.03327 \\
        beta2 & 3 & 0.02592 & 0.02451 & 0.06199 (0.00078) & \textbf{0.02398} & \textbf{0.02398} & 0.02083 \\
        beta2 & 4 & \textbf{0.01793} & 0.01874 & 0.05772 (0.00073) & 0.01952 & 0.01952 & 0.01432 \\
        beta2 & 5 & 0.01598 & \textbf{0.01470} & 0.05225 (0.00061) & 0.01513 & 0.01513 & 0.01041 \\\hline
    \end{tabular}
    \caption{Out-of-distribution (OOD) results for $n=25$, unweighted social cost. Baselines without `OOD' were trained with in-distribution (uniformly distributed) samples. Values in parentheses (if any) represent regret (max across agents) incurred by the mechanism. The best results (beside NonSP) are bolded.}
    \label{tab:res-uw-n25-ood}
\end{table*}

\clearpage
\paragraph{Generalizability to Larger Instances.}
Table \ref{tab:res-uw-n25-generalizability} shows results for $n=25$. 

\begin{table*}[h]
    \footnotesize
    \centering
    \begin{tabular}{|ccccc|ll|c|}
        \hline
        & $K$ & Per. & MoulinNet & RegretNet@25k & \multicolumn{1}{c}{LLMMech ($n=5$)} & \multicolumn{1}{c|}{LLMMech-e ($n=5$)} & NonSP \\\hline
        uniform & 1 & \textbf{0.24099} & 0.24101 & 0.24814 (0.00263) & \textbf{0.24099} & \textbf{0.24099} & 0.24099 \\
        uniform & 2 & \textbf{0.11636} & 0.11665 & 0.12503 (0.00250) & \textbf{0.11636} & \textbf{0.11636} & 0.11248 \\
        uniform & 3 & \textbf{0.07406} & 0.07456 & 0.08340 (0.00229) & 0.10088 & 0.10088 & 0.06857 \\
        uniform & 4 & \textbf{0.05294} & 0.05352 & 0.06262 (0.00228) & 0.06723 (0.00516) & 0.06723 (0.00516) & 0.04736 \\\hline
        normal & 1 & \textbf{0.08969} & 0.08977 & 0.09147 (0.00265) & \textbf{0.08969} & \textbf{0.08969} & 0.08969 \\
        normal & 2 & \textbf{0.05094} & 0.05112 & 0.05520 (0.00218) & \textbf{0.05094} & \textbf{0.05094} & 0.04904 \\
        normal & 3 & 0.03482 & \textbf{0.03473} & 0.03988 (0.00184) & 0.05035 & 0.04200 (0.00335) & 0.03213 \\
        normal & 4 & \textbf{0.02575} & 0.02606 & 0.03174 (0.00111) & 0.03404 (0.00345) & 0.03404 (0.00345) & 0.02291 \\\hline
        beta1 & 1 & \textbf{0.06439} & 0.06440 & 0.06517 (0.00199) & \textbf{0.06439} & \textbf{0.06439} & 0.06439 \\
        beta1 & 2 & \textbf{0.03440} & 0.03452 & 0.03696 (0.00296) & 0.03545 & 0.03512 & 0.03309 \\
        beta1 & 3 & 0.02279 & 0.02293 & 0.02598 (0.00277) & 0.02763 (0.00143) & \textbf{0.02237} (0.00365) & 0.02073 \\
        beta1 & 4 & \textbf{0.01622} & 0.01795 & 0.02002 (0.00296) & 0.02132 (0.00170) & 0.01858 (0.00314) & 0.01430 \\\hline
        beta2 & 1 & \textbf{0.06434} & 0.06435 & 0.06555 (0.00145) & \textbf{0.06434} & \textbf{0.06434} & 0.06434 \\
        beta2 & 2 & \textbf{0.03466} & 0.03485 & 0.03782 (0.00056) & 0.04521 & 0.03524 & 0.03327 \\
        beta2 & 3 & \textbf{0.02305} & 0.02467 & 0.02671 (0.00000) & 0.02313 (0.00149) & 0.02743 (0.00584) & 0.02083 \\
        beta2 & 4 & \textbf{0.01658} & 0.01667 & 0.02056 (0.00000) & 0.02452 & 0.01971 (0.00085) & 0.01432 \\\hline
    \end{tabular}
    \caption{Results for generalizability of LLM-generated mechanisms from $n=5$ to $n=25$, unweighted social cost. Baselines were trained directly on $n=25$ dataset.} 
    \label{tab:res-uw-n25-generalizability}
\end{table*}

\paragraph{Varying Strategyproofness Relaxation.}

\begin{figure}[h]
    \centering
    \includegraphics[width=0.45\textwidth]{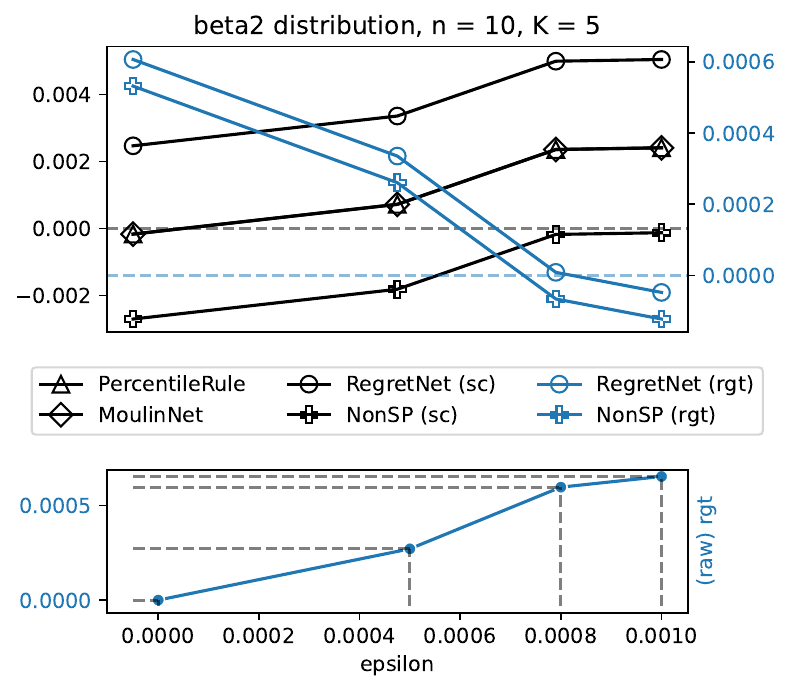}
    \caption{Summary of results on varying $\varepsilon$ of LLMMech-e for $n=10$ arbitrarily weighted agents from beta2 distribution, under $K=5$. \textbf{Top:} Reported values from black/blue lines (using left/right y-axis as reference) are net differences with respect to LLMMech-e in social cost/empirical max regret, respectively. \textbf{Bottom:} Raw values of empirical regret from LLMMech-e.} 
    \label{fig:res-aw-n10-ablation}
\end{figure}

\begin{figure*}[h]
    \centering
    \includegraphics[width=0.8\textwidth]{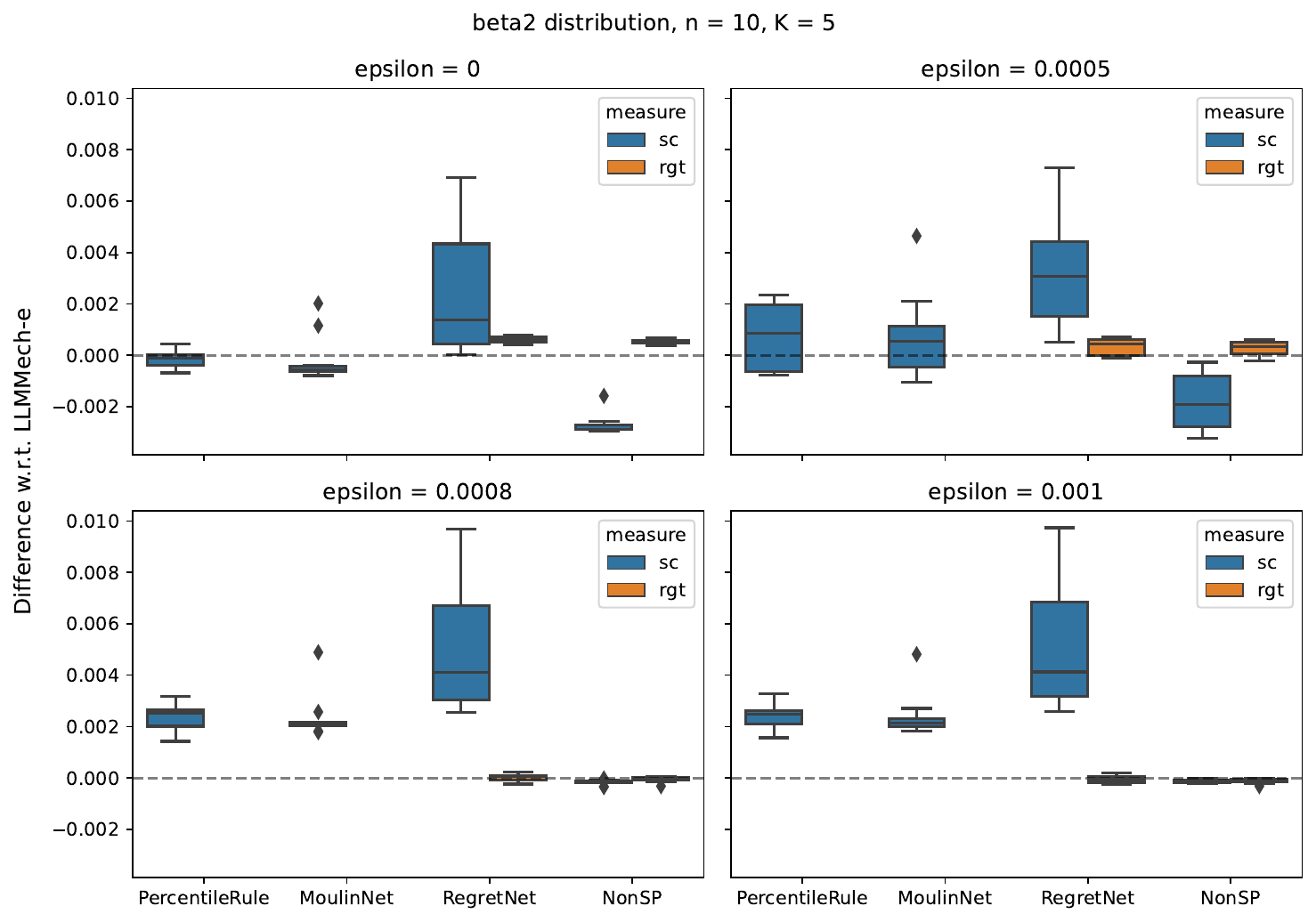}
    \caption{Complete results on varying $\varepsilon$ of LLMMech-e for $n=10$ arbitrarily weighted agents, $K=5$, beta2 distribution. We also report the (empirical) regret for RegretNet and NonSP.}
    \label{fig:res-aw-n10-ablation-full}
\end{figure*}

Following up on the previous experiments on arbitrary weights, we investigate how LLMMech-e would perform when we gradually increase $\varepsilon$ in Equation \ref{eq:fitness-penalty}. We consider beta2 distribution, where LLMMech-e seems to perform not as well compared to other distributions (Figure \ref{fig:res-aw}, last row), for $n=10$ and $K=5$.
Figure \ref{fig:res-aw-n10-ablation} summarizes our results for $\varepsilon\in\{0,0.0005,0.0008,0.001\}$. 
The top plot shows the net difference (averaged across 10 weight sets) in social cost (black lines) and in regret (blue lines) with respect to LLMMech-e for each baseline.
See Figure \ref{fig:res-aw-n10-ablation-full} for the complete results at each $\varepsilon$. 
The optimal choice of $\varepsilon$ for this particular problem configuration seems to be within $[0.0005,0,0008]$, where there is a significant improvement in social cost while maintaining empirical regret below the one from NonSP. 
\nate{We notice the empirical regret returned from the mechanisms is always (i) 0 when $\varepsilon=0$ and (ii) below $\varepsilon$ when $\varepsilon>0$ (Figure \ref{fig:res-aw-n10-ablation}, bottom). This transparent relationship means in practice, unlike RegretNet (where the only means to enforce strategyproofness is via tuning $\rho$ opaquely for each problem setting), we can either directly enforce strategyproofness or flexibly relax the strategyproofness constraint given some domain-specific threshold in terms of gain from misreporting.}

\end{document}